\documentclass[final,3p,twocolumn,authoryear]{elsarticle}

\usepackage{amssymb}

\usepackage{lineno}
\usepackage{hyperref}
\biboptions{sort&compress}
\usepackage[cmex10]{amsmath}
\usepackage{multirow, makecell} 
\usepackage{arydshln} 
\usepackage{gensymb} 
\usepackage{algorithm, algpseudocode}
\usepackage{xcolor}
\usepackage{pifont}
\newcommand{\cmark}{\ding{51}}%
\newcommand{\xmark}{\ding{55}}%
\usepackage{sistyle} 
\SIthousandsep{,}

\usepackage{changepage} 
\newlength{\offsetpage}
{\end{adjustwidth}}

\journal{Elsevier}

\let\oldequation\equation
\let\oldendequation\endequation
\renewenvironment{equation}{\linenomathNonumbers\oldequation}{\oldendequation\endlinenomath}

\newcommand{\RV}[1]{{\color{black}#1}}

\begin{document}

\begin{frontmatter}



\title{GlobalMatch: Registration of Forest Terrestrial Point Clouds by Global Matching of Relative Stem Positions}

\author[Tongji]{Xufei~Wang}
\ead{tjwangxufei@tongji.edu.cn}

\author[Tongji,TU Delft]{Zexin~Yang\corref{cor1}}
\cortext[cor1]{Corresponding author}
\ead{zexinyang@tongji.edu.cn}

\author[Tongji]{Xiaojun~Cheng\fnref{label2}}
\fntext[label1]{He unfortunately passed away before the submission of this paper.}

\author[TU Delft]{Jantien~Stoter}
\ead{j.e.stoter@tudelft.nl}

\author[ZAFU]{Wenbing~Xu}

\author[Yichun]{Zhenlun~Wu}

\author[TU Delft]{Liangliang~Nan}
\ead{liangliang.nan@tudelft.nl}

 \address[Tongji]{
 	College of Surveying and Geo-Informatics, Tongji University,
 	Shanghai,
 	200092, 
 	China}

 \address[TU Delft]{
 	3D Geoinformation Research Group, Delft University of Technology,
 	2628 BL, 
 	Delft,
 	The Netherlands}

 \address[ZAFU]{
 	College of Environment and Resources, Zhejiang A\&F University,
 	Hangzhou,
 	311300, 
 	Zhejiang,
 	China}

 \address[Yichun]{
 	Big Data Development Administration of Yichun,
 	Yichun,
 	336000, 
 	Jiangxi,
 	China}

\begin{abstract}
Registering point clouds of forest environments is an essential prerequisite for LiDAR applications in precision forestry. 
State-of-the-art methods for forest point cloud registration require the extraction of individual tree attributes, and they have an efficiency bottleneck when dealing with point clouds of real-world forests with dense trees.
We propose an automatic, robust, and efficient method for the registration of forest point clouds.
Our approach first locates tree stems from raw point clouds and then matches the stems based on their relative spatial relationship to determine the registration transformation. 
The algorithm requires no extra individual tree attributes and has quadratic complexity to the number of trees in the environment, allowing it to align point clouds of large forest environments.
Extensive experiments on forest terrestrial point clouds have revealed that our method inherits the effectiveness and robustness of the stem-based registration strategy while exceedingly increasing its efficiency.
Besides, we introduce a new benchmark dataset that complements the very few existing open datasets for the development and evaluation of registration methods for forest point clouds. 
\RV{The source code of our method and the dataset are available at \url{https://github.com/zexinyang/GlobalMatch}.}
\end{abstract}




\end{frontmatter}

\section{Introduction}
\label{introduction}
The light detection and ranging (LiDAR) technology, especially terrestrial laser scanning (TLS), has brought forest inventories to a brand new 3D era~\citep{calders2020terrestrial, disney2019terrestrial, liang2016terrestrial}. 
The registration of TLS scans collected from different viewpoints is a fundamental and necessary precondition for subsequent forestry applications.
However, the registration of forest point clouds remains a challenging task due to the mutual occlusion of forest structures.
The most common and reliable practice is marker-based registration, which requires manually placing and transporting markers and is thus expensive, laborious, and time-consuming.  
Approaches dedicated to urban scenes rely on either local~\citep{cai2019practical, huang2021pairwise} or global~\citep{chen2019plade} descriptors, which are effective for well-structured urban environments. 
However, due to the complexity, irregularity, and instability of forest environments, these methods become fragile and ineffective when applied to forest point clouds. 

Given the fact that tree stems are the most stable structure in the forest environment, existing forest point cloud registration methods rely on the extracted tree stems and other attributes for the registration. 
These methods share a similar two-stage workflow: \textit{stem mapping} followed by \textit{stem matching}~\citep{henning2006detailed, henning2008multiview, liang2013automatic, liu2017automated, kelbe2016marker, kelbe2016multiview, tremblay2018towards, dai2020fast, ge2021global, polewski2019marker, guan2019novel, hauglin2014geo, hyyppa2021efficient}.
The stem mapping stage locates tree stem positions and extracts necessary individual tree attributes, e.g., diameter at breast height (DBH) and tree height. 
The second stage establishes correspondences of tree stems across a pair of point clouds by matching their positions and attributes, which is typically conducted in an iterative trial-and-error manner (see Fig.~\ref{fig: framework comparison}(a)).
Although this stem-based strategy has been continuously improved in the past decades, significant limitations remain as follows.

\textit{1) Inefficiency in stem mapping}. 
Most methods~\citep{liang2013automatic, kelbe2016marker, kelbe2016multiview, liu2017automated, tremblay2018towards, polewski2019marker, guan2019novel, hyyppa2021efficient} adopt an existing algorithm originally developed for tree detection or parameter estimation to identify tree stems.
However, the existing stem detection algorithms~\citep{othmani2011towards, liang2011automatic, kelbe2015single, xia2015detecting, yang2016mapping, polewski2017voting, ye2020improved, hyyppa2020under} can be time-consuming since they introduce computationally intensive steps (e.g., individual tree segmentation, leaf and wood separation, and subsection stem modeling) to guarantee the completeness of detection and the high accuracy of extracted tree attributes.

\textit{2) Ambiguities and inefficiency in stem matching}.  
Most existing stem matching methods rely on both tree locations and other attributes (e.g., DBH, tree height, or tree crown)\citep{liu2017automated, kelbe2016marker, kelbe2016multiview, tremblay2018towards, dai2020fast, ge2021global, hauglin2014geo, dai2019automated}, and thus they are sensitive to ambiguities in the extracted tree attributes. 
For example, it is common that multiple trees in a plantation forest have similar DBH and height values.
Besides, the overwhelming computational burden is a fatal flaw of most stem-based matching algorithms. 
Specifically, their running time grows dramatically with the increasing number of trees due to the exhaustive nature of their verification based on the iterative trial-and-error alignment framework, which results in their inability to handle forest data containing a large number of trees, not to mention the demand for highly accurate tree attributes. 

\textit{3) Insufficient data for evaluation}. 
Though improvements have been reported, the actual progress is vague to the research community because experiments are conducted on different closed datasets.
To enable reliable evaluation and comparison of forest point cloud registration methods, open-access benchmark datasets with a large data volume and diverse forestry scenarios are urgently needed but unfortunately lacking.

In this work, we aim to resolve the above challenges to achieve efficient and robust registration of forest point clouds.
Since exploiting individual tree attributes slows down stem mapping and makes stem matching sensitive to complex forest environments, we build an efficient attribute-free registration method named \textit{GlobalMatch} for forest point clouds. 
The proposed approach relies on tree positions only, avoiding the computationally expensive and unstable tree attribute derivation process.
It consists of a stem mapping algorithm that can efficiently locate tree stems in raw scans and a simple yet significantly efficient stem matching algorithm that needs no extra tree attributes but only stem positions.
Moreover, most existing forest registration methods follow an iterative trial-and-error procedure~\citep{henning2006detailed, henning2008multiview, liang2013automatic, kelbe2016marker, kelbe2016multiview, liu2017automated, tremblay2018towards, dai2020fast, guan2020marker, ge2021global, hauglin2014geo, dai2019automated, polewski2019marker, hyyppa2021efficient} and have a complexity between cubic and exponential with regard to the number of trees~\citep{liang2013automatic, kelbe2016marker, tremblay2018towards, polewski2019marker} or keypoints~\citep{guan2020marker, dai2020fast}. 
By contrast, our work falls in the inlier-grouping framework and has quadratic time complexity, so it can be applied to point clouds of large forest environments or those with dense tree instances.
The comparison of these two underlying registration frameworks is illustrated in~Fig.~\ref{fig: framework comparison} and detailed in Section~\ref{related work stem-based registration}.

Extensive experiments show that our approach inherits effectiveness and robustness and meanwhile exceedingly increases the efficiency of the stem-based registration scheme.
For comprehensive evaluation and valid comparison of registration algorithms, we also introduce a new benchmark dataset named \textit{Tongji-Trees}, to complement the very rare publicly available data for marker-free registration of TLS scans of forest areas. In summary, our main contributions include:

\begin{itemize}
	\item [1)] A fast stem mapping method that can robustly extract stem positions from raw forest scans.
	\item [2)] An efficient, robust, and deterministic stem matching algorithm that does not require iteratively extracting the correspondences and estimating the registration transformation. 
	Our algorithm has quadratic time complexity, making it suitable for registering point clouds of large forestry areas.
	\item [3)] A benchmark dataset for evaluating forest point cloud registration methods. 
	The new dataset consists of twenty point clouds collected from four forest plots with diversity concerning understory vegetation and the density, distributions, and species of trees.
\end{itemize}

\section{Related work}
\label{related work}

In light of the extensive literature on point cloud registration~\citep{maiseli2017recent, dong2020registration}, in this section, we mainly review the works that are most relevant to ours, namely, stem mapping, stem-based registration, and benchmark datasets.

\subsection{Stem mapping}
Existing stem mapping solutions mainly fall into two categories: range image-based methods~\citep{haala2004combination, forsman20053, forsman2012estimation} and unordered point-based methods~\citep{othmani2011towards, liang2011automatic, kelbe2015single, xia2015detecting, yang2016mapping, polewski2017voting, ye2020improved, shao2020slam}. 
The former category of methods, taking advantage of the ordered image structure, can efficiently extract tree stems by grouping the pixels of the range image based on local properties. 
However, range images are not always available, which hampers their use.
The latter category of methods slices the point cloud vertically and identifies tree stems by fitting 2D circles or 3D cylinders. 
A series of contiguous circles or cylinders are then connected to form a complete tree stem. 
Typically, complex shape fitting (e.g., ellipse~\citep{ye2020improved} or tapered cylinder~\citep{kelbe2015single}), skeleton optimization~\citep{othmani2011towards}, individual tree segmentation~\citep{yang2016mapping, guan2019novel}, supervised point classification~\citep{polewski2017voting}, and growth direction constraints~\citep{xia2015detecting, ye2020improved} are used to improve detection integrity and mapping accuracy. 
However, these computationally intensive steps harm the efficiency of stem mapping.

We want to point out that existing stem mapping techniques pursue the completeness of detection results or the accuracy of calculated tree attributes using computationally expensive steps, which may not be necessary. 
In registration, not every single tree contributes to the registration. 
A good registration result can be guaranteed as long as parts of trees in the overlapping area have been identified. 
It is, therefore, valuable to design a fast stem mapping approach for forest point cloud registration.
In this work, we introduce a fast stem mapping method to locate stems from raw scans. 
As our registration method does not require accurate tree attributes, our stem mapping does not have computationally expensive steps for complete stem structure recovery.

\subsection{Stem-based registration}
\label{related work stem-based registration}

\begin{figure}[t]
    \centering
    \includegraphics[width=0.95\linewidth]{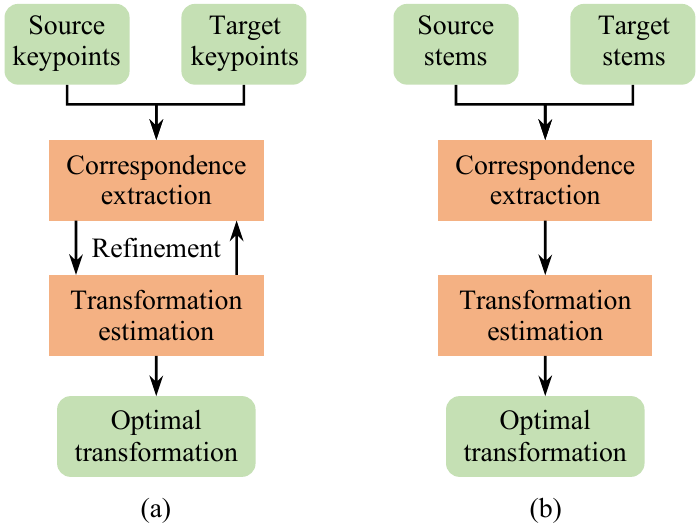}
    \caption{Comparison between the existing forest point cloud registration framework and ours. 
    (a)~The iterative trial-and-error framework. Given two sets of keypoints, correspondences and transformations are estimated and refined iteratively until the optimal transformation is found. 
    (b)~Our inlier-grouping framework. With two sets of stem positions extracted from forest scans, our method first establishes correspondences by grouping all inliers, and then it computes the optimal transformation using the correspondences. 
    Compared to the methods that follow the iterative trial-and-error framework, our method establishes correspondences and computes the transformation only once.}
    \label{fig: framework comparison}
\end{figure}

Marker-free forest registration techniques generally use tree stems as basic primitives. 
They broadly fall into two registration frameworks: iterative trial-and-error~(Fig.~\ref{fig: framework comparison}(a)) and inlier-grouping~(Fig.~\ref{fig: framework comparison}(b)). 

\subsubsection{Iterative trial-and-error based methods}
In these methods~\citep{henning2006detailed, henning2008multiview, liang2013automatic, kelbe2016marker, kelbe2016multiview, liu2017automated, tremblay2018towards, dai2020fast, guan2020marker, ge2021global, hauglin2014geo, dai2019automated, polewski2019marker, hyyppa2021efficient}, finding the optimal transformation is achieved through an iterative process.
First, a minimum sample set (MSS) is chosen to calculate a transformation. 
Using this transformation, source stems are then transformed to the coordinate system of target stems. 
Finally, correspondences are extracted by matching close trees with similar tree attributes and employed to re-estimate the transformation. 
The above three steps are iteratively conducted until the optimal transformation (i.e., the one with the largest number of corresponding trees) is found.
One well-known paradigm of this type of method is the random sample consensus (RANSAC) algorithm~\citep{fischler1981random, kelbe2015single, tremblay2018towards}.

These methods are robust but computationally expensive since, in each iteration, all the stem positions are transformed and used to determine the inliers.
Attempting all possible MSSs is thus not computationally affordable.
For example, restricted by the heavy computational burden of examining all possible stem matches, \citet{liang2013automatic} only manage to achieve 2D horizontal alignment of forest TLS scans and remain the vertical translation unsolved.
On the other hand, sampling MSS randomly (i.e., RANSAC) does not perform well at a high outlier ratio (which is common in forest environments) since its running time increases exponentially with the outlier ratio~\citep{cai2019practical, kelbe2015single}.
To enable the practical registration of point clouds of large forest environments, current studies have proposed three strategies to address the computational inefficiency.

The first strategy is to discard unlikely MSSs. 
Specifically, the similarities derived from relative stem positions~\citep{hauglin2014geo, kelbe2016marker, kelbe2016multiview, liu2017automated, tremblay2018towards, hyyppa2021efficient} and additional tree attributes (e.g., DBH~\citep{hauglin2014geo, kelbe2016marker, kelbe2016multiview, liu2017automated, tremblay2018towards} and tree height~\citep{hauglin2014geo, liu2017automated}) are used to eliminate false MSSs and sort the remaining ones. 
These methods work well for forests of small or medium sizes. 
However, most of them remain inefficient when dealing with large-scale forests. 
For example, the memory usage and execution time of~\citet{kelbe2016marker} can reach prohibitively large when the forest scene has more than 50 trees, while its efficient version~\citep{tremblay2018towards} also takes more than an hour for registering scans containing more than 100 trees.
Moreover, they can become impractical in plantation forests due to the ambiguities of tree attributes.
Using only the locations of trees as input, the 2D coarse registration method presented by \citet{hyyppa2021efficient} can efficiently co-register multiplatform LiDAR data in large-scale forests.
However, the algorithm is designed to solve a 2D registration problem and only works with well-leveled point clouds (i.e., the Z axis of the point cloud points upward).

The second strategy is to reduce the number of keypoints. 
Recent studies introduce novel keypoints, e.g., visual occlusion points identified from shaded areas~\citep{guan2020marker} or mode points extracted from tree crowns~\citep{dai2019automated, dai2020fast}. 
Compared to stems, these local keypoints typically have a smaller number. They are used to register forest scans~\citep{dai2019automated, guan2020marker} or to guide the stem matching~\citep{dai2020fast} to bypass the computational burden of matching numerous stems. 
Notwithstanding the enhanced efficiency, these local keypoints require high-quality input point clouds that are still challenging to acquire for complex forest environments~\citep{ge2021global}. 
Furthermore, similar to stem positions, the increase in the number of keypoints also results in a dramatic increase in computation time~\citep{guan2020marker}.

The third strategy is to use advanced optimization techniques.
\citet{ge2021global} register multiview forest point clouds by adopting the 4-points congruent sets algorithm~\citep{aiger20084, ge2017automatic} that reduces the number of trials required to establish a reliable registration.
\citet{polewski2019marker} apply simulated annealing metaheuristic~\citep{van1987simulated} to determine the optimal corresponding trees between multiplatform LiDAR point clouds.
\citet{dai2019automated} consider the alignment as a maximum likelihood estimation problem and employ the coherent point drift algorithm~\citep{myronenko2010point} to fuse forest airborne and terrestrial point clouds.

\subsubsection{Inlier-grouping based methods}
This category of methods seeks the optimal transformation in a one-shot manner.
Specifically, all corresponding trees (i.e., inliers) are first extracted and then employed to determine the optimal transformation.
On the one hand, rather than verifying the transformation and refining correspondences in each iteration as in the iterative trial-and-error framework, inlier-grouping methods extract correspondences and estimate the transformation only once and are thus generally more efficient~\citep{zhao2021comprehensive}.
On the other hand, as the transformation is estimated only once, the resulting transformation may not be accurate if the correspondences are unreliable~\citep{gressin2013towards, dai2019automated}. 
Therefore, the key to these methods is to extract true correspondences (i.e., inliers) and eliminate false ones (i.e., outliers)~\citep{cai2019practical}.

Though the inlier-grouping framework has been widely adopted in registration techniques targeting man-made objects~\citep{chen20073d, albarelli2010game, aldoma2012global} or urban scenes~\citep{yang2016urban, cai2019practical}, it is rarely used in forest point cloud registration.
To the best of our knowledge, the co-registration approach for multiplatform forest LiDAR data proposed by~\citet{guan2019novel} is the only one that falls into this framework.
For each stem position extracted from raw scans via individual tree segmentation, the horizontal coordinates and neighboring stem positions are used to construct a 2D triangulated irregular network (TIN). 
Then, stem correspondences are extracted by matching 2D TINs (where the matching quality is measured by the number of similar triangles within the TINs) and employed to determine the registration transformation.
Their method can reach a satisfactory accuracy (i.e., less than 20 cm) in co-registering multiplatform LiDAR scans, while their running time has not been reported.

In this work, we propose a stem matching method that follows the inlier-grouping framework.
We generalize the urban matching approach of~\citet{yang2016urban} to forest environments, and we tailor three main improvements for robust and efficient forest point cloud registration.
First, we utilize only geometric coordinates because the semantic features defined in their method are not available in forest environments.
Second, by constructing local triangles that encode relative stem positions, we reduce the algorithmic complexity concerning the number of trees from sextic to quadratic.
Last, unlike their greedy grouping algorithm that directly adopts the line-to-line geometric consistency of~\citet{chen20073d}, we propose a graph-to-graph consistency measure that guarantees the robustness in triangle-based correspondence grouping.

\subsection{Benchmark datasets}
Existing benchmark datasets for point cloud registration typically target man-made objects, indoor scenes, or urban environments~\citep{mian2006three, theiler2015globally, sanchez2017global, chen2019plade, dong2020registration}.
Very few datasets are publicly available for the evaluation of registration methods for forest point clouds.
To the best of our knowledge, WHU-FGI~\citep{dong2020registration} and ETH-Trees~\citep{theiler2015globally} are the only two datasets for this purpose, which contain only five and six scans of a single forest plot, respectively.
The other FGI forest dataset~\citep{liang2018international} is intended for assessing tree attribute extraction algorithms, which cannot be used for evaluating registration methods due to the inaccessibility of the registration ground truth.
In this work, we release Tongji-Trees, a new benchmark dataset consisting of twenty scans for the comprehensive evaluation of marker-free registration methods for TLS scans of forest areas.

\section{Methodology}
\label{methodology}

\begin{figure*}[t]
	\centering
	\includegraphics[width=0.95\linewidth]{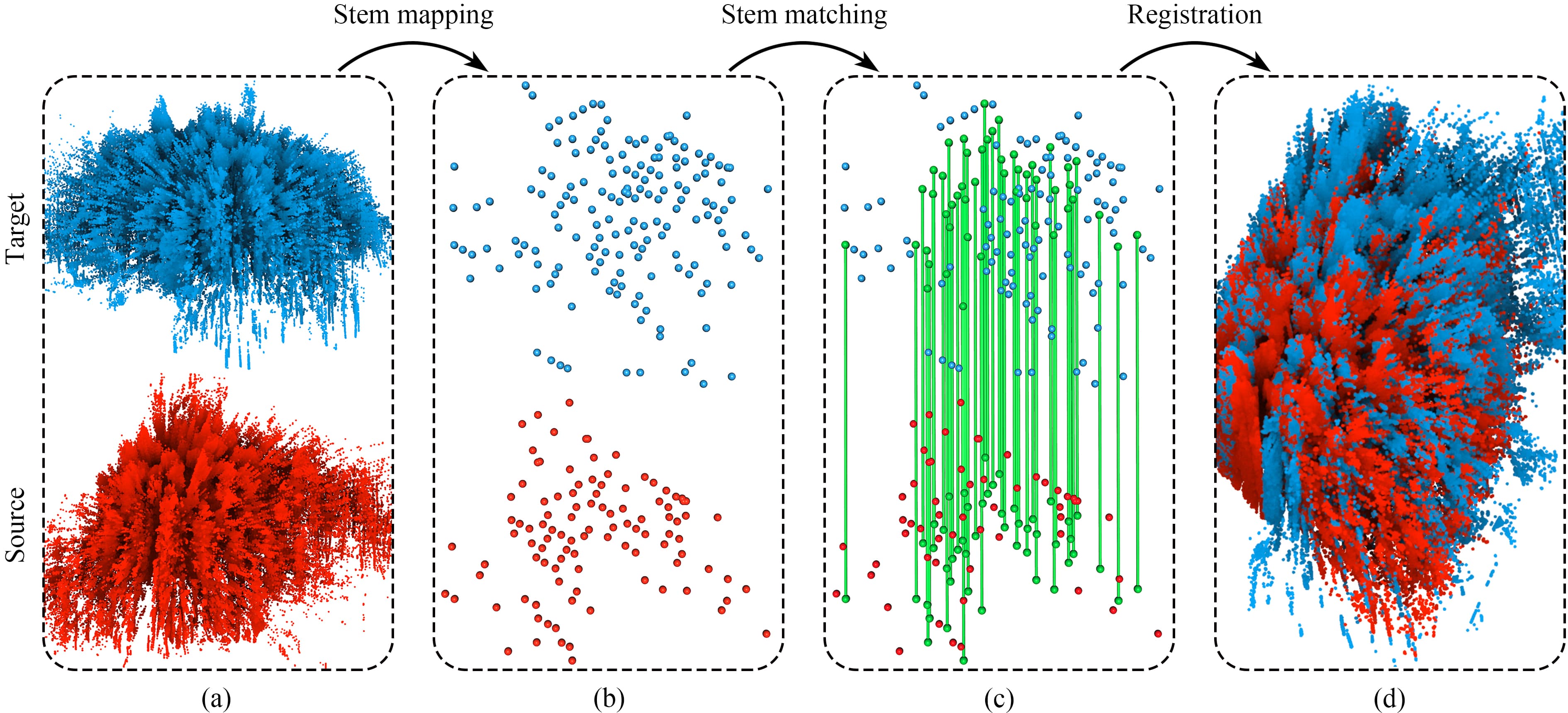}
	\caption{The pipeline of the proposed method.
	    (a)~Input point clouds.
	    (b)~Stem positions.
	    (c)~Stem matches.
	    (d)~Registration result. 
		Given a pair of forest TLS point clouds as input~(a), the stem positions of trees are first extracted from each point cloud~(b).
		Then, correspondences between stem positions are established by exploiting relative distances of the stem positions~(c).
		Finally, registration is achieved by applying the transformation computed from the stem correspondences~(d).}
	\label{fig: pipeline}
\end{figure*}

Our method takes as input a pair of forest TLS scans and outputs a rigid transformation to register the two scans. 
It consists of three main stages illustrated in Fig.~\ref{fig: pipeline}: stem mapping, stem matching, and registration, which are detailed as follows.

\subsection{Stem mapping}
\label{subsection stem mapping}
This stage aims to identify stem positions, i.e., the intersections of stem axes and the ground surface, from a raw TLS scan.
This is achieved by first extracting stems, followed by intersecting the stem axes and the ground surface.

\begin{figure}[t]
    \centering
    \includegraphics[width=0.95\linewidth]{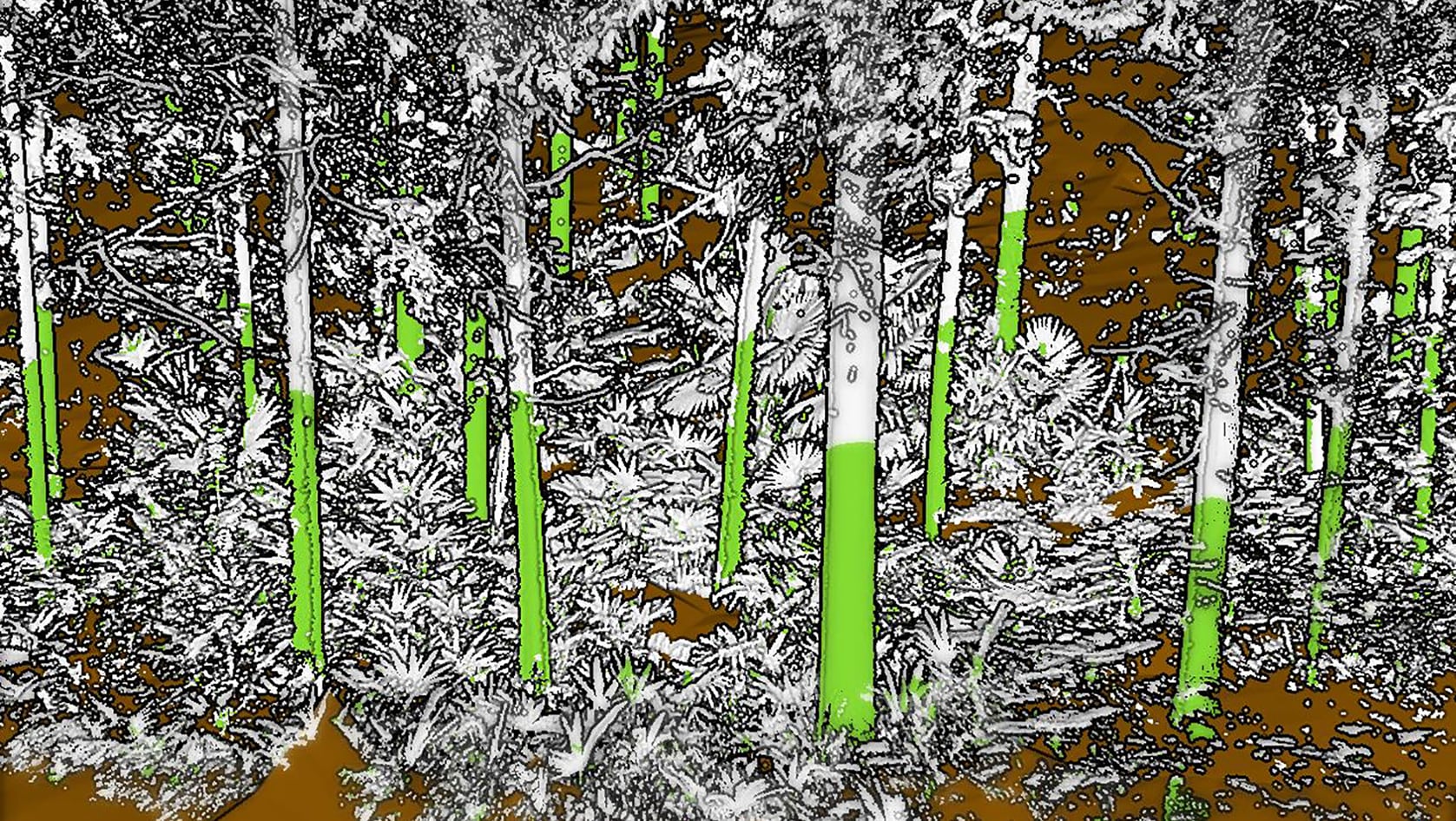}
    \caption{An example of stem extraction results (a close-up view). The reconstructed DTM is visualized in brown and the extracted tree stems are in green. In this point cloud, the number of points is 30,574,461, while the number of extracted tree stems is only 116. Our method uses the extracted tree stems for registration.}
    \label{fig: stem regions}
\end{figure}

To avoid interference from unstable structures such as forest canopy and understory vegetation, we identify the stem points of trees before the extraction of individual stems. 
To filter out forest canopies, we create a digital terrain model (DTM) based on the input point cloud~\citep{zhang2016easy} and mark the points within 0.2--3~m from the DTM as the understory layer.
Before excluding understory shrubs, we downsample the data of the understory layer for efficiency reasons. We exploit a voxel grid structure with a resolution of 1 cm, and for each voxel, only the point that is closest to its centroid is kept.
Considering that tree stems typically have a vertical orientation, we identify tree stems by looking into the pointwise verticality~\citep{demantke2012streamed, weinmann2013feature} defined as 
\begin{equation}
\label{equation: verticality}
v = 1 - \left| n_z \right|, 
\end{equation}
where $n_z$ represents the third component of the normal vector of a point. A value of 1 indicates that the local surface of a point is perfectly vertically oriented, while a zero value indicates a perfectly horizontal local geometry. 
In this work, we use a fixed radius of 10~cm for querying the neighbors of a point using the method described in~\citet{behley2015efficient} to estimate its normal using principal component analysis~\citep{sanchez2020robust}.
We mark the points whose verticality is higher than a threshold $\gamma$ as stem points.
Because most real-world trees are near vertical, we empirically set $\gamma$ to 0.9 in all our experiments. 
With this threshold, stem points are identified as those whose normal deviates within 5.8 degrees from the horizontal direction. 
A smaller $\gamma$ is recommended if the majority of trees in the data are inclined.
By extracting the stem points, we also significantly reduce the amount of data to be further processed. 
Fig.~\ref{fig: stem regions} shows an example of the stem extraction results, in which the number of points in the extracted stems is only about $1/80$ the size of the initial input. 

With the stem points identified from the previous step, we extract individual tree stems by detecting cylinders using the RANSAC algorithm~\citep{fischler1981random}. 
Since detecting multiple cylinders by sequentially applying RANSAC is inefficient and sub-optimal (e.g., a previous wrong detection usually harms the subsequent detection)~\citep{pham2016geometrically}, we first employ the Euclidean clustering algorithm~\citep{rusu2009laser} to separate the stem points into small groups each containing a single stem (or in very rare cases few adjacent stems for trees with multi-branch structures near the ground surface). 
Then for the points of each stem, we apply RANSAC to detect a cylinder. 
It is worth noting that a few incorrectly detected stems will not affect the subsequent registration because they do not have correspondences in the other point cloud and are thus filtered out in the stem matching stage.
Finally, we obtain a set of stem positions $\mathcal{L} = \{ \mathbf{l}_i, 1 \leq i \leq N_l \}$ by intersecting the axes of the cylinders with the DTM model, where $N_l$ denotes the number of stem positions.

\subsection{Stem matching}
\label{correspondence matching}

\begin{figure*}[t]
    \centering
    \includegraphics[width=0.95\linewidth]{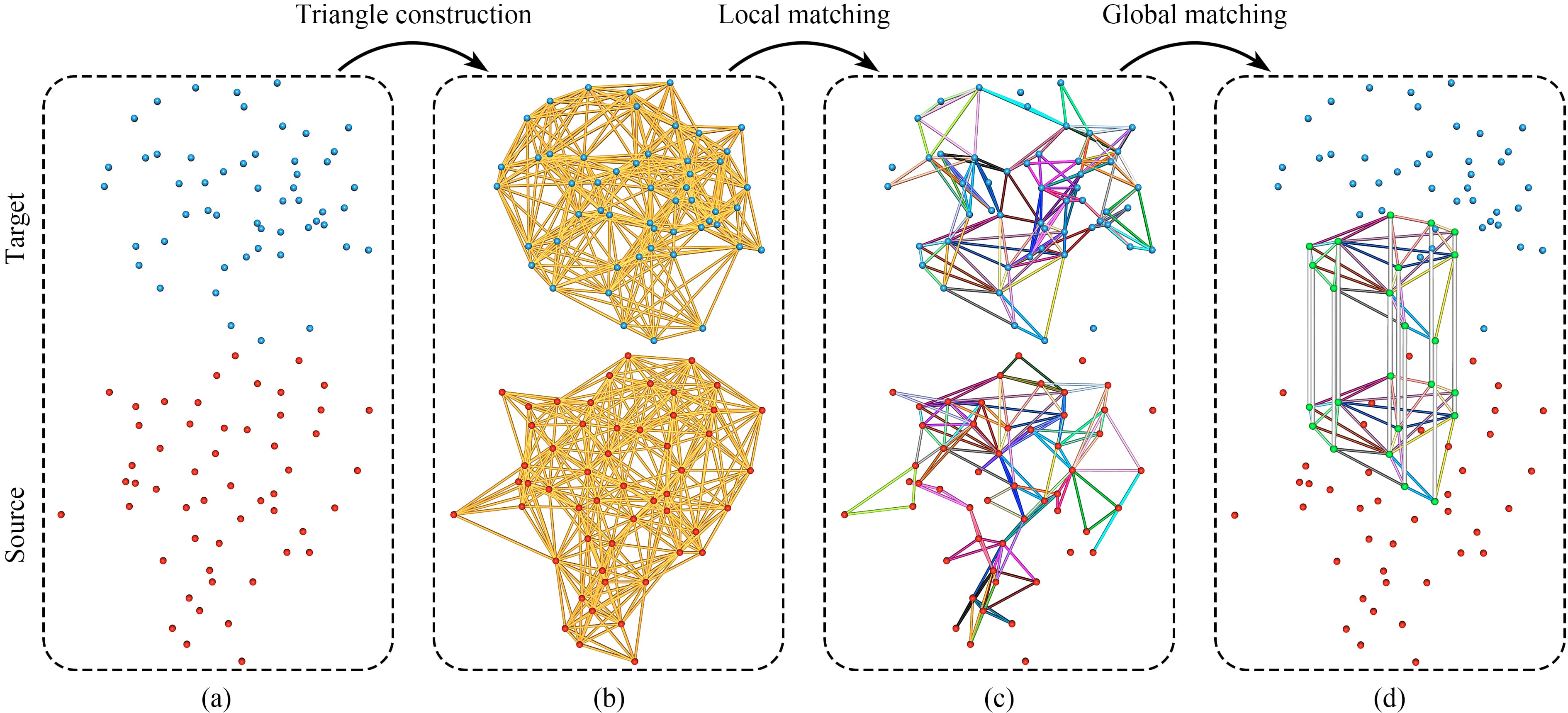}
    \caption{Stem matching.
        (a)~Stem positions. 
        (b)~Triangles encoding the local relative relationship of stem positions. 
        (c)~Locally matched triangle pairs. Each pair is assigned a random color for the visualization purpose.
        (d)~Globally matched triangle pairs and stem matches.
        In this example, 12 stem matches (denoted by the white lines in~(d)) are extracted from 53 target stem positions (denoted by the blue dots in~(a)) and 53 source stem positions (denoted by the red dots in~(a)).
        The numbers of initial triangles for the target and source scans are both 523~(b). The numbers of locally and globally matched triangle pairs are 43~(c) and 13~(d), respectively.}
    \label{fig: matching}
\end{figure*}

The goal of this stage is to establish the correspondences between the previously extracted stem positions of the two input point clouds. 
Our stem matching method differs from most existing methods in the following two aspects: 
(1)~it requires only the information of stem positions while existing methods~\citep{liu2017automated, kelbe2016marker, kelbe2016multiview, hauglin2014geo, tremblay2018towards, dai2020fast, ge2021global} also require additional tree attributes, such as DBH, tree height, or tree crown; 
(2)~our stem matching is a deterministic non-iterative process while existing methods~\citep{henning2006detailed, henning2008multiview, liang2013automatic, kelbe2016marker, kelbe2016multiview, liu2017automated, tremblay2018towards, dai2020fast, guan2020marker, ge2021global, hauglin2014geo, dai2019automated, polewski2019marker, hyyppa2021efficient} follow an iterative trial-and-error procedure that is less efficient.

Our stem matching consists of three steps as shown in Fig.~\ref{fig: matching}. 
We first construct a set of local triangles for each scan from its stem positions to encode their relative spatial relationship. 
These triangles are then matched in the subsequent local matching and global matching steps to obtain the correspondences between the stem positions of the two input scans.
In the following, we elaborate on each step of our stem matching algorithm and derive its overall time complexity.

\subsubsection{Triangle construction}
RANSAC-based methods~\citep{kelbe2016marker, kelbe2016multiview, tremblay2018towards, dai2020fast} commonly use an exhaustive approach to create triangles from the stem positions, i.e., any triplet of stem positions forms a triangle.
Thus, the number of triangle pairs to be verified using this exhaustive approach is 

\begin{figure}[t]
    \centering
    \includegraphics[width=0.95\linewidth]{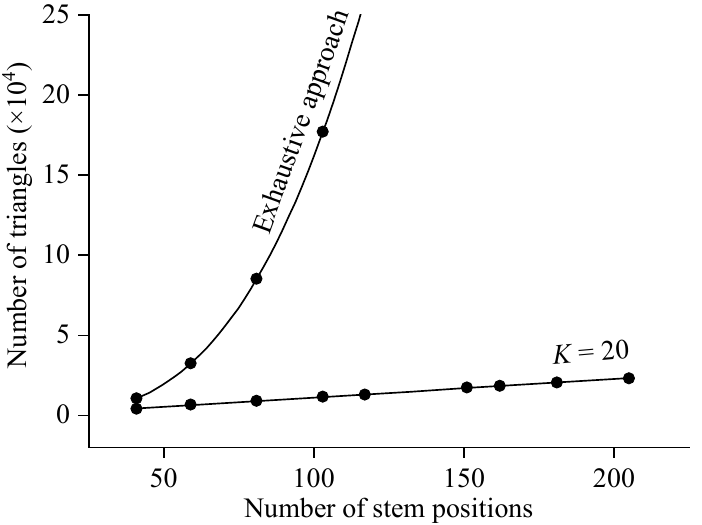}
    \caption{
    The relationship between the number of constructed triangles and the number of stem positions.
    Curves for the exhaustive approach and our method (with $K = 20$) are plotted using stem positions extracted from the proposed Tongji-Trees dataset.
    For our method, the number of triangles increases linearly with respect to the number of stem positions, while the increase for the exhaustive approach is cubic.}
    \label{fig: computational complexity}
\end{figure}

\vspace{-0.5cm}
\RV{$$\binom{N_l^\star}{3} \cdot \binom{N_l^\prime}{3},$$}
where \RV{$N_l^\star$} and \RV{$N_l^\prime$} denote the number of stems in the source and target scans, respectively.
With this exhaustive approach, the number of triangles to be matched increases explosively with the number of stems extracted from the input point clouds. 
We observe that only certain triangles whose vertices have correspondences in the other point cloud contribute to stem matching and it is sufficient to build triangles by only connecting the adjacent stem positions. 
To this end, we construct only local triangles by connecting every stem position with its $K$-nearest stem positions. 
This simple strategy significantly reduces the complexity of triangle matching from \RV{$\mathcal{O}\left( (N_l^\star)^3 \cdot (N_l^\prime)^3 \right)$} to \RV{$\mathcal{O}\left( N_l^\star \cdot N_l^\prime \right)$}.
Specifically, with an increasing number of detected stem positions in a scan, the number of triangles created by the exhaustive approach increases cubically, i.e.,
\begin{equation}
\label{eq: cubic number}
N_e = \frac{1}{6}N_l^3-\frac{1}{2}N_l^2+\frac{1}{3}N_l,
\end{equation}
\noindent while that of our method increases linearly, i.e., 
\begin{equation}
\label{eq: linear number}
N_K = \frac{K^2-K}{2} N_l,
\end{equation}
\noindent where $N_l$ is the number of stem positions, and $N_e$ and $N_K$ denote the number of triangles constructed by the exhaustive approach and our method, respectively.
In our implementation, we set $K$ to $20$, which ensures sufficient local triangles to be constructed to establish the correspondences between the two sets of stem positions. 
Fig.~\ref{fig: computational complexity} depicts how the number of triangles increases for both methods. 
\RV{Our triangle construction step has quasilinear complexity of $\mathcal{O}\left( N_l^\star \log N_l^\star + N_l^\prime \log N_l^\prime \right)$.}

With the set of triangles $\mathcal{T}^\star=\{t_i^\star, 1 \leq i \leq N_t^\star \}$ constructed from the source scan and the set of triangles $\mathcal{T}^\prime=\{t_i^\prime, 1 \leq i \leq N_t^\prime \}$ constructed from the target scan, where $N_t^\star$ and $N_t^\prime$ denote the number of triangles in the source and target scans respectively, we establish the correspondences between the source and target triangles in two steps: a local matching step and a global matching step.

\subsubsection{Local matching}
In this step, we establish triangle-wise correspondences between the two sets of triangles. 
This can be easily achieved by comparing the edge lengths of two triangles from the two sets.
For efficiency reasons, we first sort the vertices of each triangle such that the one opposite to the longest edge comes first and the triangle is in counter-clockwise orientation.
The following lemma lays the foundation for our triangle-wise local matching.

\newtheorem{lemma}{Lemma}
\begin{lemma}
\label{lemma1}
If three stem positions in the source scan $\mathbf{l}_i^\star,\mathbf{l}_j^\star,\mathbf{l}_k^\star \in \mathcal{L}^\star$ have correspondences in the target scan
$\mathbf{l}_i^\prime,\mathbf{l}_j^\prime,\mathbf{l}_k^\prime\in \mathcal{L}^\prime$, the source triangle $\triangle l_i^\star l_j^\star l_k^\star \in \mathcal{T}^\star$ is congruent with the target triangle $\triangle l_i^\prime l_j^\prime l_k^\prime \in \mathcal{T}^\prime$. 
\end{lemma}


\noindent As the contrapositive of Lemma~\ref{lemma1}, the following lemma is equally true: 

\begin{lemma}
\label{lemma2}
Given three pairs of stem positions $\langle \mathbf{l}_i^\star,\mathbf{l}_i^\prime \rangle$, $\langle \mathbf{l}_j^\star,\mathbf{l}_j^\prime \rangle$, $\langle \mathbf{l}_k^\star, \mathbf{l}_k^\prime \rangle$, if the source triangle $\triangle l_i^\star l_j^\star l_k^\star \in \mathcal{T}^\star$ is not congruent with the target triangle $\triangle l_i^\prime l_j^\prime l_k^\prime \in \mathcal{T}^\prime$, then at least one pair of the stem positions does not match. 
\end{lemma}

By verifying any pair of source and target triangles, we accumulate potentially matched triangle pairs (Lemma~\ref{lemma1}) and filter out false matches (Lemma~\ref{lemma2}).  
In our work, we define the local dissimilarity between two triangles $t_1$ and $t_2$ as 
\begin{equation}
\label{local consistency}
D_{local}(t_1, t_2) = \sum_{i=1}^3 \left| \varepsilon^\star_i - \varepsilon^\prime_i \right|, \forall \left| \varepsilon^\star_i - \varepsilon^\prime_i \right| < \epsilon,
\end{equation}
where $\varepsilon^\star_i$ and $\varepsilon^\prime_i$ denote respectively the length of the corresponding edges in the two triangles. 
$\epsilon$ is the maximum allowed difference in edge length for two triangles to be matched, which is empirically set to 5~cm in all our experiments.
Note that if the difference in edge length of any pair of edges is greater than $\epsilon$, the two triangles are not considered matched. 
For each triangle $t_j^\prime$ in the target scan, we find its potential corresponding triangle $t_i^\star$ in the source scan as the one that has the minimum dissimilarity value, resulting in a locally matched triangle pair $u = \langle t_i^\star, t_j^\prime \rangle$.
The local matching step has quadratic complexity of $\mathcal{O}\left( N_t^\star \cdot N_t^\prime \right)$, where $N_t^\star$ and $N_t^\prime$ are the number of triangles constructed in the source and target scans, respectively.

After local matching, a considerable portion of triangles for which local matches could not be found will be deleted. 
Since the locally matched triangle pairs may contain outliers, as seen in Fig.~\ref{fig: matching}(c), a global matching step is necessary because two triangles being congruent is only a necessary (but not sufficient) condition for the stem positions to be correspondent.

\subsubsection{Global matching}

This step aims to establish the correspondences between the two sets of stem positions encoded in the source triangles and target triangles respectively. 
In theory, this can be achieved by matching two undirected fully connected graphs whose vertices are the stem positions of the source and target scans, respectively. 
We approach the graph matching by looking for the largest consensus set of the two graphs.

Unlike the local matching step that compares individual triangles, our global matching step uses triangle pairs as primitives. 
Before explaining our global matching, we first define a metric called global dissimilarity to measure the consistency between two pairs of locally matched triangles. 

\begin{figure}[t]
    \centering
    \includegraphics[width=0.95\linewidth]{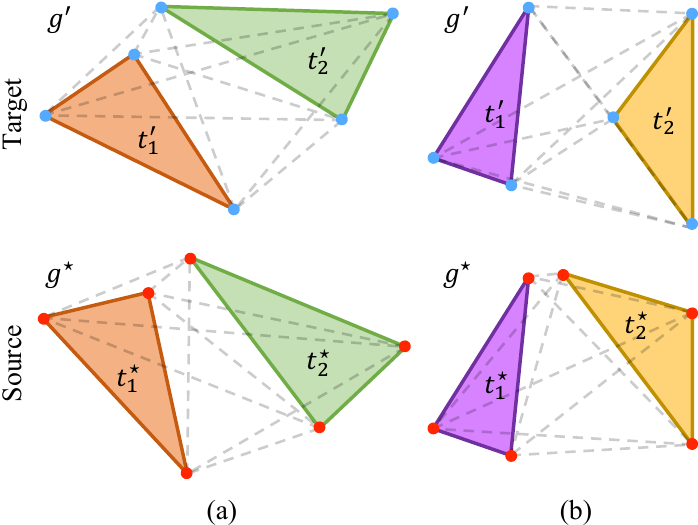}
    \caption{Two examples illustrating the global dissimilarity between two locally matched triangle pairs $u_1 = \langle t_1^\star, t_1^\prime \rangle$ and $u_2 = \langle t_2^\star, t_2^\prime \rangle$. 
    (a)~$u_1$ and $u_2$ form a consensus set, i.e., $D_{global} (u_1, u_2) < \epsilon$.
    (b)~$u_2$ does not agree with $u_1$ (i.e., $D_{global} (u_1, u_2) > \epsilon$) and thus they cannot form a consensus set.}
    \label{fig: global constraint}
\end{figure}

Given two locally matched triangle pairs $u_1 = \langle t_1^\star, t_1^\prime \rangle$ and $u_2 = \langle t_2^\star, t_2^\prime \rangle$ (see Fig.~\ref{fig: global constraint}), we build two undirected fully connected graphs, i.e., $g^\star$ from the vertices of $t_1^\star$ and $t_2^\star$, and $g^\prime$ from the vertices of $t_1^\prime$ and $t_2^\prime$. 
Then the global dissimilarity between the two locally matched triangle pairs $u_1$ and $u_2$ is defined as
\begin{equation}
\label{global consistency}
D_{global} (u_1, u_2) = \max \left| \mathcal{E}^\star_i - \mathcal{E}^\prime_i \right|,
\end{equation}
where $\mathcal{E}^\star_i$ and $\mathcal{E}^\prime_i$ denote the lengths of the corresponding edges respectively in $g^\star$ and $g^\prime$. 
Since $t_1^\star$ and $t_1^\prime$, $t_2^\star$ and $t_2^\prime$ are respective congruent triangle pairs, it is sufficient to test only the edges connecting vertices of different triangles (i.e., the edges denoted by the dashed lines in Fig.~\ref{fig: global constraint}). 
${D}_{global}$ measures how much adding a new pair of locally matched triangles disagrees with an existing pair. 
If a locally matched triangle pair is consistent with any existing pair, we call it a globally matched pair, just to differentiate it from the locally matched pairs.
Note that~\citet{yang2016urban} compare only the relative positions of the centroids of two triangles, which has ambiguities. 
See Fig.~\ref{fig: global constraint}(b) for an example, where the distance between the centroids of $t_1^\prime$ and $t_2^\prime$ is identical to that of $t_1^\star$ and $t_2^\star$ but the two pairs of triangles do not match.

\begin{figure*}[t]
    \centering
    \includegraphics[width=0.95\linewidth]{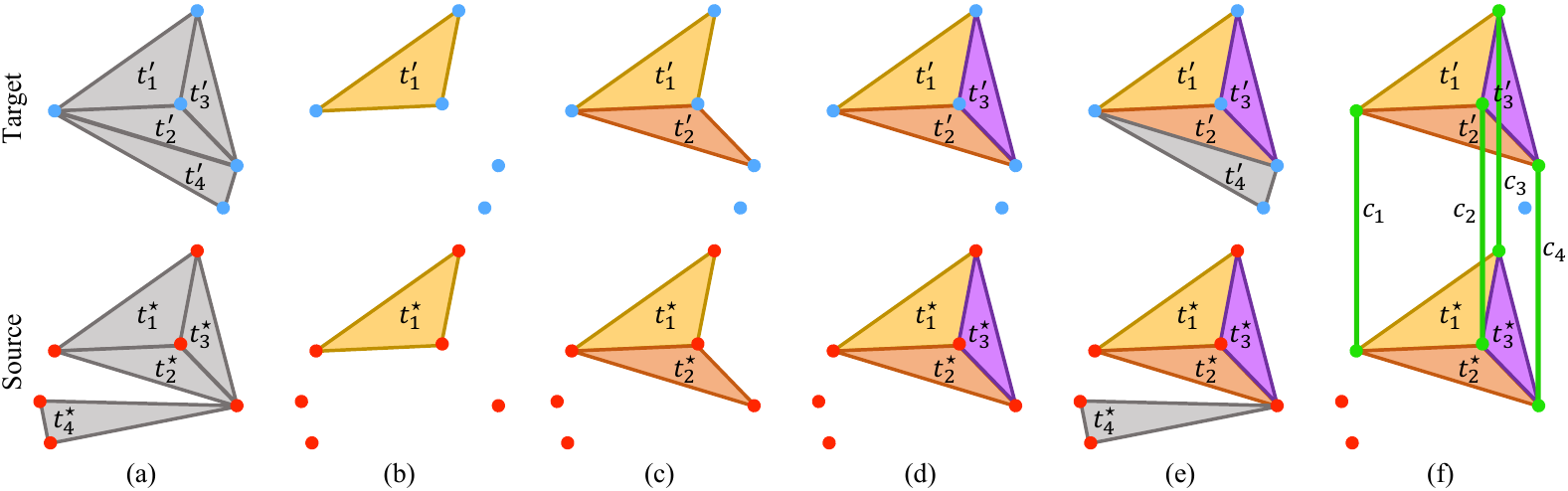}
    \caption{
    An example illustrating the growth of a consensus set.
        Given an initial set of locally matched triangle pairs $\mathcal{U} = \{ u_1, u_2, u_3, u_4 \}$, $u_i = \langle t_i^\star, t_i^\prime \rangle$~(a), our algorithm starts accumulating globally matched triangle pairs from an initial pair $\langle t_1^\star, t_1^\prime \rangle$~(b).
        In each iteration, our method tries to find one more globally matched triangle pair and adds it to the growing set~((c) and (d)). 
        The growing process stops when no more triangle pair that is consistent with the initial pair can be found. 
        In this example, the triangle pair $u_4$ does not agree with the initial pair $u_1$ due to $D_{global} (u_1, u_4) > \epsilon$~(e).   
        Finally, a set of globally matched triangle pairs $\mathcal{U}^{\dag} = \{ u_1, u_2, u_3 \}$ is obtained, giving us the correspondences of the stem positions $\mathcal{C} = \{ c_1, c_2, c_3, c_4 \}$~(f). 
        The correspondences are denoted by the green lines connecting the stem positions of the two scans.}
    \label{fig: global matching}
\end{figure*}

We obtain the optimal correspondences between the two sets of stem positions by finding the largest consensus set. For each locally matched triangle pair $u_i$, our method grows a consensus set by accumulating globally matched triangle pairs $u_j$ that agree with $u_i$ (i.e., $D_{global} (u_i, u_j) < \epsilon$, where $\epsilon$ is the same threshold used in the local matching step). Fig.~\ref{fig: global matching} illustrates the growth of a consensus set of globally matched triangle pairs. 
Each consensus set grows independently and the one that has accumulated the most globally matched triangle pairs gives us the optimal correspondences between the two sets of stem positions. The process for finding the largest consensus set is detailed in Algorithm~\ref{alg: global matching}. 

The global matching step has quadratic complexity of $\mathcal{O}\left( N_u \cdot (N_u-1) \right)$, where $N_u$ represents the number of locally matched triangle pairs.
It is worth noting that the growth of a consensus set can be immediately stopped when a sufficient number of globally matched triangle pairs have been accumulated.
However, the growing process only involves comparing edge lengths and it is quite efficient to find the largest consensus set using an exhaustive search because of the small number of triangle pairs that remained after the local matching step. 
Such an exhaustive search guarantees the optimal consensus set is always used for the registration, ensuring robustness in registration. 
Note that our global matching step is carried out over the already matched local triangle pairs, and it has quadratic complexity (measured against the number of locally matched triangle pairs).
Existing methods, such as~\citep{kelbe2016marker, kelbe2016multiview, tremblay2018towards, dai2020fast}, perform an exhaustive search on the initial sets of stem positions, which has sextic time complexity (measured against the number of stem positions).
Thus, the complexity of our global matching is several orders of magnitude smaller. 
For example, for the scan pairs shown in Fig.~\ref{fig: matching}, our global matching does 2756 times triangle matching test, while the methods~\citep{kelbe2016marker, kelbe2016multiview, tremblay2018towards} will have to evaluate \num{548777476} times.

\subsubsection{Overall time complexity}
Before deriving the overall time complexity of the proposed stem matching algorithm, notations for quantities are first recalled for readability.
$N_l$, $N_t$, and $N_u$ are the number of detected stem positions, constructed triangles, and locally matched triangle pairs, respectively.
The superscript $\star$ represents the source scan, while the superscript $\prime$ denotes the target scan.

The overall time complexity of our method can be derived as follows.
We can consider the majority of input stem positions are inliers, i.e., $N_l^\star \approx N_l^\prime \approx N_l$. 
In this case, every triangle can find a local match in the other scan. Thus, the number of locally matched triangle pairs $N_u \approx N_t^\star \approx N_t^\prime \approx N_t$, which leads to the worst-case time complexity.
Because of the linear relationship between $N_t$ and $N_l$ (see Eq.~\ref{eq: linear number}), the overall complexity can be given by \RV{$\mathcal{O}\left( 2 \cdot N_l \log N_l + N_l \cdot N_l + N_l \cdot (N_l - 1) \right)$}, which can be further simplified to $\mathcal{O}\left( N_l^2 \right)$. 
Therefore, our method has quadratic time complexity.

\begin{algorithm}[t]
\small
  \caption{Global matching}
  \label{alg: global matching}
  \renewcommand{\algorithmicrequire}{\textbf{Input:}}
  \renewcommand{\algorithmicensure}{\textbf{Output:}}
  \begin{algorithmic}[1]
  \Require a set of locally matched triangle pairs $\mathcal{U}$
  \Ensure correspondences of stem positions $\mathcal{C}$
  \State \textbf{Initialization:} $\mathcal{C} \leftarrow \varnothing$; each consensus set $\mathcal{U}_{i}^{\dag} \leftarrow \varnothing$
  \For{each $u_i \in \mathcal{U}$}
    \State insert $u_i$ into $\mathcal{U}_{i}^{\dag}$
    \For{each $u_j \in \mathcal{U}, i \neq j$}
        \If{$D_{global}(u_i, u_j) < \epsilon$}
          \State insert $u_j$ into $\mathcal{U}_{i}^{\dag}$
        \EndIf
      \EndFor
  \EndFor
  \State $\mathcal{U}_{max}^{\dag} \leftarrow$ $\max( \{ \mathcal{U}_{i}^{\dag} \} )$
  \For{each pair of corresponding vertices $c_i \in \mathcal{U}_{max}^{\dag}$} 
    \State insert $c_i$ into $\mathcal{C}$ 
  \EndFor
  \end{algorithmic}
\end{algorithm}

\subsection{Registration}
\label{method_registration}
From the correspondences of stem positions, a rigid transformation can be computed to register the two input scans.
As there is no scale change among TLS scans, this transformation has 6 degrees of freedom (DoFs) and can be written as a $4 \times 4$ matrix with the following structure
\begin{equation}
\mathbf{T} = 
\left[
    \begin{matrix}
       \begin{array}{c : c}
            \mathbf{R} & \mathbf{t} \\ \hdashline
            \mathbf{0} & 1
        \end{array} 
    \end{matrix}
\right]
=
\left[
    \begin{matrix}
        \begin{array}{c c c : c}
            R_{11} & R_{12} & R_{13} & t_x \\
            R_{21} & R_{22} & R_{23} & t_y \\
            R_{31} & R_{32} & R_{33} & t_z \\ \hdashline
            0      & 0      & 0      & 1
        \end{array} 
    \end{matrix}
\right],
\end{equation}
\noindent where $\mathbf{R}$ is the 3-DoF rotation and $\mathbf{t}$ is the 3-DoF translation.
More information about computing a rigid transformation from a set of correspondences can be found in~\citet{sorkine2017least} and~\citet{paulj1992method}.

Using the above computed transformation, we can register the two input point clouds into the same coordinate system.
Considering that scanners nowadays typically produce well-leveled point clouds, the complexity of the registration problem for such data can be reduced to 4-DoF, i.e., a translation (3 DoFs) and a rotation around the Z axis (1 DoF), and the final registration transformation has the following structure,
\begin{equation}
\mathbf{\dot{T}} = 
\left[
\begin{matrix}
\begin{array}{c : c}
\mathbf{R}(\phi) & \mathbf{t} \\ \hdashline
\mathbf{0}            & 1
\end{array} 
\end{matrix}
\right]
=
\left[
\begin{matrix}
\begin{array}{c c c : c}
\cos{\phi} & -\sin{\phi} & 0 & t_x \\
\sin{\phi} & \cos{\phi}  & 0 & t_y \\
0          & 0           & 1 & t_z \\ \hdashline
0          & 0           & 0 & 1
\end{array} 
\end{matrix}
\right],
\end{equation}
where $\mathbf{R}(\phi)$ defines the 1-DoF rotation about the azimuth $\phi$. 
In this case, the 4-DoF registration can be achieved by the alignment of the two scans in the 2D horizontal direction (3~DoFs) and a 1D vertical translation (1~DoF). 
Specifically, with the $x$ and $y$ coordinates of the corresponding stem positions, we estimate the 2D horizontal registration parameters $\phi$, $t_x$, and  $t_y$ by solving a least-squares problem \citep{sorkine2017least}.
Using the $z$ coordinates of the corresponding stem positions $\langle z_i^\star, z_i^\prime \rangle$, we calculate the vertical translation $t_z=\frac{1}{N_c}\sum_{i=1}^{N_c} \left( z_i^\prime - z_i^\star \right)$, where $N_c$ is the number of correspondences of stem positions.

Our method allows both 6-DoF registration and 4-DoF registration.
When aligning forest TLS scans that are well leveled, the 4-DoF solution is preferred since it takes advantage of the scanners and may achieve better coarse registration accuracy. 
We provide an experimental comparison between our 6-DoF and 4-DoF registration strategies in Section~\ref{comparison of 6-DoF and 4-DoF}.
For the remaining experiments, we employed the 4-DoF solution.

After the coarse registration of the two scans, a fine registration algorithm can be employed to further improve the registration accuracy. 
Although there are fine registration algorithms (e.g.,~\cite{shao2020slam}) developed for forest scenarios, we use the iterative closest points (ICP) algorithm~\citep{paulj1992method} because of its simplicity and effectiveness.

\subsection{Implementation details}
For efficiency, we parallelize computationally intensive parts of the proposed algorithm.
Specifically, the following parts of our algorithm are parallelized:
\begin{itemize}
	\item neighborhood query, verticality calculation, cylinder detection, and stem position computation in the stem mapping stage;
	\item local matching and global matching in the stem matching stage.
\end{itemize}

\section{Benchmark dataset}
\label{benchmark dataset}
    
\subsection{Study Areas}
\begin{figure*}[t]
    \centering
    \includegraphics[width=0.95\linewidth]{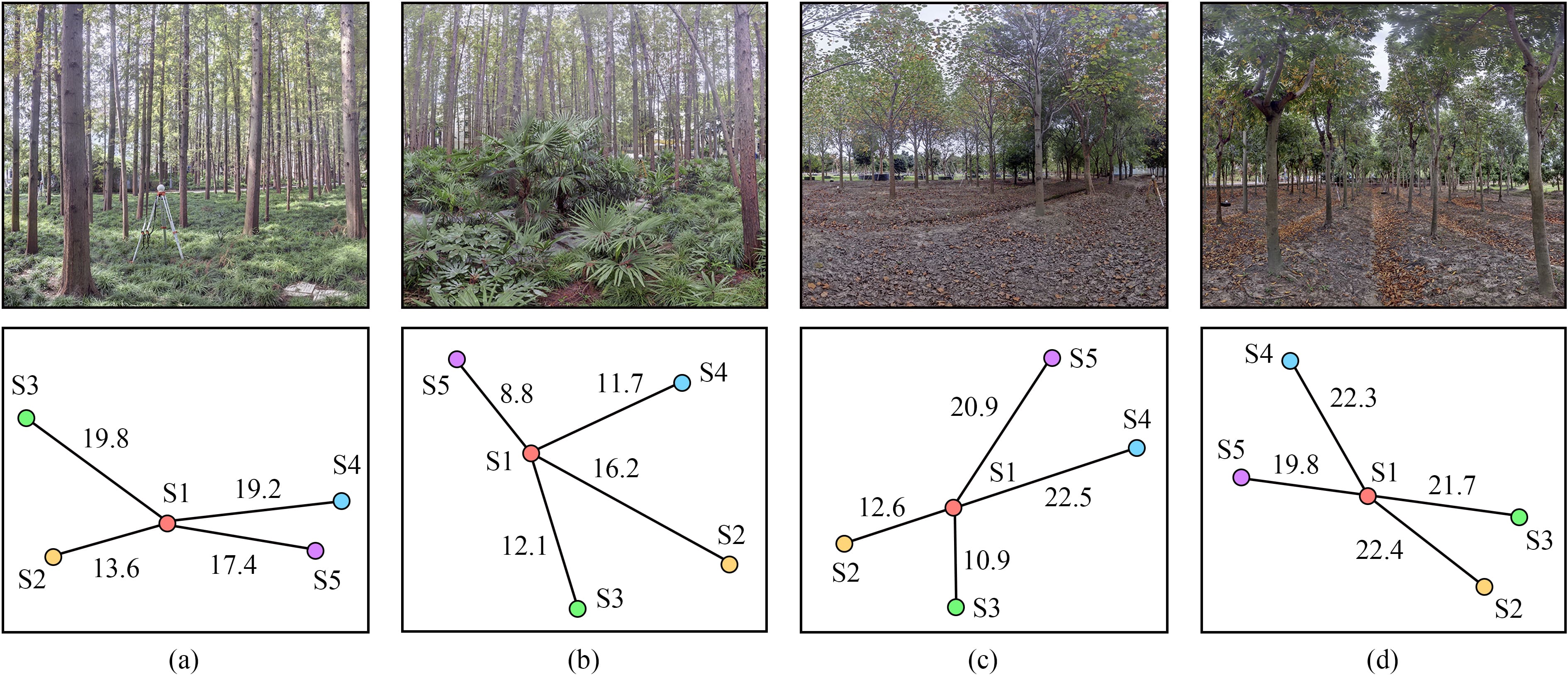}
    \caption{The four plots of the Tongji-Trees dataset.
    (a)~Plot~\#1. 
    (b)~Plot~\#2. 
    (c)~Plot~\#3. 
    (d)~Plot~\#4. 
    The top row shows a representative view and the bottom row shows the five scanning positions (S1--S5) for each plot. 
    The numbers indicate the distances (in meters) between the related scanning positions.}
    \label{fig: study areas}
\end{figure*}

The Tongji-Trees dataset contains TLS scans of four plantation forest plots, which were collected in Shanghai, China (31.15\degree N, 121.12\degree E). 
The plot size varies from Plots~\#1 to \#4, with a cover area of $65 \times 50$, $90 \times 50$, $80 \times 40$, and $65 \times 55$~m$^2$, respectively. 
Metasequoia trees are irregularly distributed in Plots~\#1 and \#2, with an average height of 19~m and 23~m and a density of 1036~trees/ha and 1210~trees/ha, respectively. 
As for Plots~\#3 and \#4, Liriodendron chinense trees with an average height of 7~m and Sapindus mukorossi trees with an average height of 10~m were planted in rows with spacing of 5~m and~3 m, respectively. 
Compared to Plots~\#1 and \#2, Plots~\#3 and \#4 are sparser, with a density of 433~trees/ha and 982~trees/ha, respectively. 
As shown in the top row of Fig.~\ref{fig: study areas}, the surface of Plot~\#1 is covered with wild grass. 
Plot~\#2 is partially overgrown with breast-high shrubs.
Both Plots~\#3 and \#4 have bare but humped surfaces. 
The mutual occlusion of complex shrubs in Plot~\#2 makes it more challenging for registration than in the other plots.
Table~\ref{table: dataset} summarizes the information of all the plots in the Tongji-Trees dataset.

\begin{table*}[t]
\footnotesize
\centering
\caption{Statistics on the Tongji-Trees benchmark dataset \label{table: dataset}}
\setlength{\leftskip}{-30pt}
\setlength{\tabcolsep}{3.5 pt}
\begin{tabular}{c c c c c c c c c c}
\hline
\multirow{2}{*}{\textbf{Plot ID}} &
  \multirow{2}{*}{\textbf{Main tree species}} &
  \textbf{Plot size} &
  \textbf{Tree height} &
  \textbf{Tree density} &
  \multicolumn{5}{c}{\textbf{Number of points of each scan}} \\
       &                       & (m$^2$)        & (m) & (trees/ha)   & S1         & S2         & S3         & S4         & S5         \\
\hline
\#1   & Metasequoia            & 65 $\times$ 50 & 19  & 1036         & \num{33540068} & \num{33469682} & \num{13553220} & \num{34488581} & \num{33652969} \\
\#2   & Metasequoia            & 90 $\times$ 50 & 23  & 1210         & \num{33661875} & \num{33891812} & \num{30574461} & \num{34193448} & \num{33197039} \\
\#3   & Liriodendron chinense  & 80 $\times$ 40 & 7   & 433          & \num{35510140} & \num{35169411} & \num{33680749} & \num{33445603} & \num{37076208} \\
\#4   & Sapindus mukorossi     & 65 $\times$ 55 & 10  & 982          & \num{37189070} & \num{35274954} & \num{37113902} & \num{35598180} & \num{35783768} \\
\hline
\end{tabular}
\end{table*}

\subsection{Data acquisition}
By exploiting the ``multi-scan'' data acquisition approach~\citep{liang2016terrestrial}, each of the four plots was scanned at five positions, i.e., one at the center and the other four away around the center, forming five scans for each plot and twenty scans for the whole dataset. 
The bottom row of Fig.~\ref{fig: study areas} illustrates the scanner positions for each plot in the Tongji-Trees dataset. 
During field surveys, we used a Z+F~5010C terrestrial laser scanner to capture the point cloud data. 
The scanner has a measurement accuracy of $\pm$3~mm at 50~m, a measurement range of 0.3--180~m, and a view angle of $360^{\circ} \times 320^{\circ}$. 

\subsection{Ground truth}
Before scanning, we strictly leveled the scanner and set up spherical reference markers that could help us to produce correspondences for deriving the ground-truth registration transformation.
Each marker was mounted on a tripod or placed stably on the ground (rather than fixed on tree stems, which usually causes occlusions).
Each raw scan contains about 30 million points (see Table~\ref{table: dataset}), with each point represented as a 7D vector ($xyz$ coordinates, $rgb$ colors, and an intensity value). 
Although our registration method relies on 3D coordinates only, the provided attributes may benefit other methods or studies. 
We identified the markers and served them as correspondences to compute the ground-truth transformation matrix for every possible pair of scans.
The ground-truth transformation matrices have an average registration accuracy of 2.8~mm and thus can be used to evaluate the performance of registration methods.

\section{Results and discussion}
\label{experiments}

\subsection{Experimental setup}
Our algorithm is implemented in C++ using Point Cloud Library \citep{rusu20113d} and OpenMP \citep{dagum1998openmp}. 
All experiments were carried out on a laptop with an Intel Core 2.60 GHz i7 CPU (8 threads) and 32~GB RAM.

\subsubsection{Test datasets}
We have tested the proposed method on our Tongji-Trees dataset, the WHU-FGI dataset~\citep{dong2020registration}, and the ETH-Trees dataset~\citep{theiler2015globally}.
The WHU-FGI dataset contains five scans captured from a $32 \times 32$~m$^2$ plot of a natural forest dominated by Scots pines, located at Evo, Finland (61.19\degree N, 25.11\degree E).
The ETH-Trees dataset consists of six scans acquired in a forest with a large amount of underwood.

\subsubsection{Evaluation metrics}
\RV{
To provide a comprehensive evaluation, we evaluated separately the results of coarse registration and fine registration (i.e., using the ICP algorithm~\citep{paulj1992method} as a post-process).
We have recorded both matrix-based errors and pointwise errors.
Furthermore, we have calculated the success rate of coarse registration methods.
}

\begin{itemize}
	\item Matrix-based errors, i.e., the rotation error $e_R$ and the translation error $e_t$ between the estimated transformation parameters $\mathbf{R}, \mathbf{t}$ and the corresponding ground-truth values $\mathbf{\tilde{R}}, \mathbf{\tilde{t}}$: 
	\begin{equation}
	e_R = \arccos \left( \frac{\mathrm{tr}(\mathbf{\tilde{R}} \mathbf{R}^\intercal) - 1}{2} \right),
	\end{equation}
	\begin{equation}
	e_t = \left\|\mathbf{t} - \mathbf{\tilde{t}}\right\|,
	\end{equation}
	where $e_R$ measures the difference between the two rotation matrices $\mathbf{R}$ and $\mathbf{\tilde{R}}$, and $e_t$ quantifies the magnitude of the difference of the two translation vectors $\mathbf{t}$ and  $\mathbf{\tilde{t}}$.
	It is worth noting that the matrix-based metric is not straightforward for comparing two registration methods because a method may perform better in one measure but worse in the other (see Plot \#1 in Table~\ref{table: coarse results}). 
	In this case, it is not obvious to determine which method is more effective.
	
	\item Pointwise error, proposed by~\citet{chen2019plade}.
	Given a source scan $\mathcal{P}^\star=\{\mathbf{p}_i^\star, 1 \leq i \leq N_p^\star \}$ containing $N_p^\star$ points, the pointwise error is defined as
	\begin{equation}
	\label{pointwise error}
	e_p = \frac{1}{N_p^\star}\sum_{i=1}^{N_p^\star}\left\|\mathbf{R}\mathbf{p}_i^\star + \mathbf{t} -\left(\mathbf{\tilde{R}}\mathbf{p}_i^\star + \mathbf{\tilde{t}}\right)\right\|.
	\end{equation}
	Compared to the matrix-based errors, the pointwise error is more intuitive in comparing the performance of different registration methods.
	
	\item Success rate.
	A coarse registration is considered successful if the registered point cloud is sufficiently close to the ground truth, i.e., with a pointwise error less than a given threshold. 
	This threshold was set to 50~cm for our test datasets following the evaluation method described in~\citet{chen2019plade}. 
	The success rate of coarse registration is measured by the number of successful registrations divided by the total number of scan pairs.
\end{itemize}

\RV{
Matrix-based and pointwise errors are determined primarily by the quality of point clouds and the consistency of keypoints extracted from them.
These metrics are only meaningful if the algorithm finds the correct correspondences.
A failed registration will result in random and extremely large error values ($>$ meters).
Furthermore, successful coarse results can be optimized via the ICP algorithm to achieve similarly satisfactory accuracy (see Table~\ref{table: coarse results}).
However, when a coarse registration fails, the fine-registration step becomes infeasible, resulting in complete failure in pairwise registration.
From this point of view, we consider the success rate as the main metric for evaluating coarse registration.
}

\subsection{Results}

\subsubsection{Stem mapping results}
We quantitatively evaluated the effectiveness of our stem mapping method. 
Using the manually identified stem positions as ground truth, we calculated the precision, recall, and F1-score of our automatic detection results. 
As can be seen from Table~\ref{table: mapping evaluation}, our stem mapping approach achieved 95.1\%, 95.4\%, and 95.2\% in terms of average precision, recall, and F1-score, respectively on all six test plots.

\begin{table*}[t]
\footnotesize
\caption{Quantitative evaluation of stem mapping results on the three test datasets, i.e., Tongji-Trees~(Plots~\#1--\#4), WHU-FGI~\citep{dong2020registration}, ETH-Trees~\citep{theiler2015globally}}
\label{table: mapping evaluation}
\centering
\setlength{\tabcolsep}{10 pt}
\begin{tabular}{c c c c c c c c}
\hline
\multirow{2}{*}{\textbf{Metric}} & \multicolumn{6}{c}{\textbf{Results of stem mapping}~(\%)} \\
                                  & Plot~\#1 & Plot~\#2 & Plot~\#3 & Plot~\#4 & WHU-FGI & ETH-Trees & Average \\
\hline
Precision                         & 96.0     & 97.6     & 96.1     & 92.9     & 97.1    & 90.8      & 95.1 \\
Recall                            & 96.8     & 96.2     & 94.6     & 95.5     & 95.8    & 93.4      & 95.4 \\
F1-score                          & 96.4     & 96.9     & 95.4     & 94.2     & 96.4    & 92.1      & 95.2 \\
\hline
\end{tabular}
\end{table*}

\begin{table*}[t]
\footnotesize
\caption{
Accuracy of the proposed method on the three test datasets, i.e., Tongji-Trees~(Plots~\#1--\#4), WHU-FGI~\citep{dong2020registration}, ETH-Trees~\citep{theiler2015globally}.
$e_p$ denotes the pointwise error, while $e_R$ and $e_t$ represent the two components (i.e., rotation and translation errors) of matrix-based errors, respectively.}
\label{table: fine results}
\centering
\setlength{\tabcolsep}{10 pt}
\begin{tabular}{c c c c c c c}
\hline
\multirow{2}{*}{\textbf{Plot ID}} &
  \multicolumn{2}{c}{\textbf{Forest environment}} &
  \multicolumn{4}{c}{\textbf{Fine registration errors}} \\
  & Tree locations  & Understory  & $e_p$ (cm)   &    & $e_R$ (mrad)   & $e_t$ (cm)  \\
\hline
Plot~\#1  & Dense, scattered  & Wild grass  & 0.7 &  & 0.5 & 0.7 \\
Plot~\#2  & Dense, scattered  & High shrub  & 1.3 &  & 1.2 & 1.0 \\
Plot~\#3  & Sparse, regular   & Bare ground & 0.9 &  & 0.3 & 0.8 \\
Plot~\#4  & Dense, regular    & Bare ground & 1.0 &  & 0.3 & 1.0 \\
WHU-FGI   & Sparse, scattered & Underwood   & 1.6 &  & 0.9 & 1.4 \\
ETH-Trees & Sparse, scattered & Underwood   & 0.8 &  & 0.9 & 0.6 \\
\hline
Average & \multicolumn{2}{c}{}              & 1.0 &  & 0.7 & 0.9 \\
\hline
\end{tabular}
\end{table*}

\subsubsection{Registration results}

Following the previous studies~\citep{liang2013automatic, liu2017automated, tremblay2018towards, guan2020marker, dai2020fast}, we aligned scans at the corners of each plot to the one at the center. This is a common practice for registering point clouds of LiDAR-based forest inventory.

\begin{figure*}[htbp]
    \centering
    \includegraphics[width=0.95\linewidth]{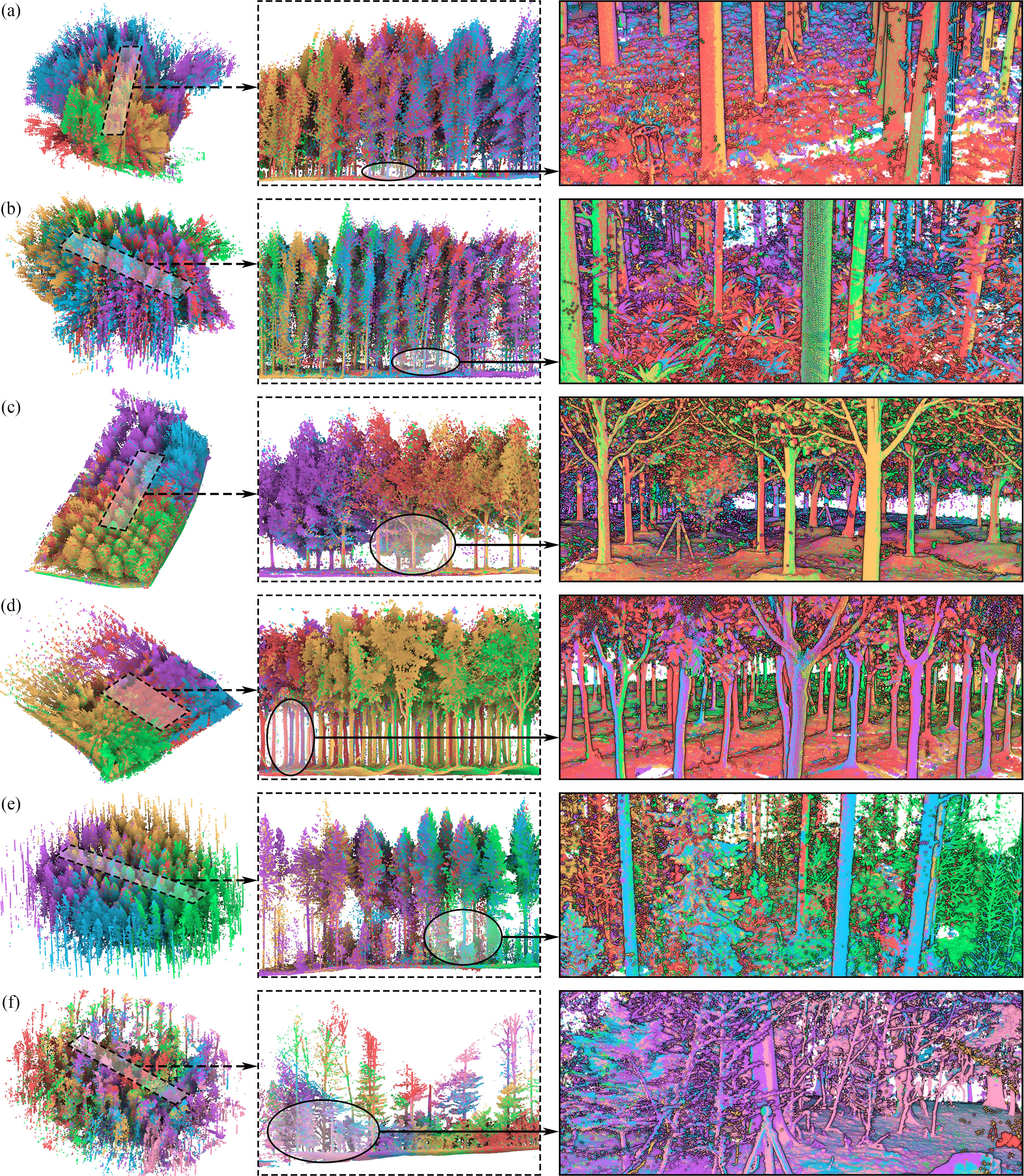}
    \caption{Registration results on the three test datasets.
    (a) to (d)~Plots~\#1--\#4 of the Tongji-Trees dataset, respectively. 
    (e)~WHU-FGI~\citep{dong2020registration}. (f)~ETH-Trees~\citep{theiler2015globally}. 
    From left to right: an overview, a cross-section, and a close-up view of the registered point clouds of each plot.
    Scans S1--S5 in each plot are visualized in red, yellow, green, blue, and purple colors, respectively, and the S6 of ETH-Trees is in pink.}
    \label{fig: registration results}
\end{figure*}

Fig.~\ref{fig: registration results} demonstrates the final registration results of our proposed method on the three test datasets.
As can be seen from the close-up view of the registration result of each plot, our method properly registered all TLS scans from diverse forest environments.
The test datasets contain both sparse forests (with a tree density below 500~trees/ha, as shown in Fig.~\ref{fig: registration results}(c) and (e)) and dense forests (with a tree density around 1000~trees/ha, as shown in Fig.~\ref{fig: registration results}(a), (b), and (d)).
Since our method has quadratic complexity to the number of trees, it can register the scans of both sparse and dense forests.
Compared to plots covered in wild grass (see Fig.~\ref{fig: registration results}(a)) and plots having bare ground (see Fig.~\ref{fig: registration results}(c) and (d)), Plot~\#2 is partially overgrown with breast-high shrubs (see Fig.~\ref{fig: registration results}(b)), whereas the WHU-FGI and ETH-Trees datasets contain a large amount of underwood.
Despite the complex understory structures, our method still extracted valid stem positions and managed to register these point clouds.
Our method is also robust to the distribution of trees. 
It correctly aligned scans containing scattered trees (Fig.~\ref{fig: registration results}(a), (b), (e), and (f)) and scans consisting of trees planted in rows (Fig.~\ref{fig: registration results}(c) and (d)).
Besides, our registration method succeeded on different tree species, i.e., Metasequoia (Fig.~\ref{fig: registration results}(a) and (b)), Liriodendron chinense (Fig.~\ref{fig: registration results}(c)), Sapindus mukorossi (Fig.~\ref{fig: registration results}(d)), and Scots pines (Fig.~\ref{fig: registration results}(e)).

Table~\ref{table: fine results} reports a quantitative evaluation of our final registration results.
The average pointwise error of all test plots is 1~cm, showing the great potential of the proposed approach to substitute the costly marker-based methods.

\begin{table*}[t]
\footnotesize
\caption{
Accuracy comparison with three competing coarse registration algorithms, i.e., TOA~\citep{kelbe2016marker}, Fast-TOA~\citep{tremblay2018towards}, and FMP+BnB~\citep{cai2019practical}.
\RV{Every method successfully registered all scan pairs from the side scans of each plot to the central ones.
All coarse registration results were further optimized by using the ICP algorithm~\citep{paulj1992method} as a fine registration step.}
$e_p$ denotes the pointwise error, while $e_R$ and $e_t$ represent the two components (i.e., rotation and translation errors) of matrix-based errors, respectively.}
\label{table: coarse results}
\setlength{\tabcolsep}{6.0 pt}
\centering
\begin{tabular}{c c c c c c c c c}
\hline
\multirow{2}{*}{\textbf{Plot ID}} &
  \multirow{2}{*}{\textbf{Method}} &
  \multicolumn{3}{c}{\textbf{Coarse registration errors}} & & \multicolumn{3}{c}{\textbf{Fine registration errors}} \\
                           &          & $e_p$ (cm)   & $e_R$ (mrad)  & $e_t$ (cm) & & $e_p$ (cm)   & $e_R$ (mrad)  & $e_t$ (cm) \\
\hline
\multirow{4}{*}{Plot \#1}  & TOA      & 11.0         & 10.9      & 8.7     &  & 0.7     & 0.5       & 0.7     \\
                           & Fast-TOA & 9.4          & 9.2       & 6.7     &  & 0.7     & 0.5       & 0.7     \\
                           & FMP+BnB  & 9.1          & 3.2       & 8.8     &  & 0.7     & 0.6       & 0.7     \\
                           & Ours     & \textbf{6.2} & 0.7       & 6.3     &  & 0.7     & 0.5       & 0.7     \\
\hline
\multirow{4}{*}{Plot \#2}  & TOA      & 27.8         & 22.4      & 21.4    &  & 1.2     & 1.2       & 0.9     \\
                           & Fast-TOA & 27.3         & 24.7      & 19.7    &  & 1.5     & 1.2       & 1.2     \\
                           & FMP+BnB  & 6.6          & 2.5       & 6.3     &  & 1.1     & 1.2       & 0.9     \\
                           & Ours     & \textbf{6.2} & 1.2       & 6.0     &  & 1.3     & 1.2       & 1.0     \\
\hline
\multirow{4}{*}{Plot \#3}  & TOA      & 1.4          & 1.5       & 1.3     &  & 0.7     & 0.3       & 0.8     \\
                           & Fast-TOA & \textbf{1.3} & 1.3       & 1.2     &  & 0.7     & 0.3       & 0.7     \\
                           & FMP+BnB  & 3.5          & 1.1       & 3.5     &  & 1.1     & 0.4       & 1.0     \\
                           & Ours     & 3.6          & 0.5       & 3.6     &  & 0.9     & 0.3       & 0.8     \\
\hline
\multirow{4}{*}{Plot \#4}  & TOA      & 3.9          & 3.6       & 3.8     &  & 0.9     & 0.3       & 0.9     \\
                           & Fast-TOA & 3.7          & 3.6       & 3.5     &  & 0.9     & 0.3       & 0.9     \\
                           & FMP+BnB  & 6.3          & 1.8       & 6.3     &  & 1.1     & 0.4       & 1.1     \\
                           & Ours     & \textbf{3.1} & 0.3       & 3.1     &  & 1.0     & 0.3       & 1.0     \\
\hline
\multirow{4}{*}{WHU-FGI}   & TOA      & 7.6          & 7.6       & 5.5     &  & 1.7     & 1.0       & 1.5     \\
                           & Fast-TOA & \textbf{6.1} & 7.5       & 3.7     &  & 1.7     & 1.0       & 1.5     \\
                           & FMP+BnB  & 7.3          & 3.6       & 6.9     &  & 1.4     & 0.8       & 1.2     \\
                           & Ours     & 7.6          & 0.6       & 7.6     &  & 1.6     & 0.9       & 1.4     \\
\hline
\multirow{4}{*}{ETH-Trees} & TOA      & 20.7         & 22.8      & 18.2    &  & 0.8     & 0.9       & 0.6     \\
                           & Fast-TOA & 22.2         & 20.0      & 20.0    &  & 0.8     & 0.9       & 0.7     \\
                           & FMP+BnB  & \textbf{5.0} & 3.6       & 4.5     &  & 0.8     & 0.9       & 0.6     \\
                           & Ours     & 8.7          & 3.4       & 8.6     &  & 0.8     & 0.9       & 0.6     \\
\hline
\multirow{4}{*}{Average}   & TOA      & 12.1         & 11.4      & 9.8     &  & 1.0     & 0.7       & 0.9     \\
                           & Fast-TOA & 11.7         & 11.0      & 9.1     &  & 1.0     & 0.7       & 1.0     \\
                           & FMP+BnB  & 6.3          & 2.6       & 6.1     &  & 1.0     & 0.7       & 0.9     \\
                           & Ours     & \textbf{5.9} & 1.1       & 5.8     &  & 1.0     & 0.7       & 0.9     \\
\hline
\end{tabular}
\end{table*}

\subsection{Comparison}
We have compared our method with three existing coarse registration algorithms: TOA~\citep{kelbe2016marker}, Fast-TOA~\citep{tremblay2018towards}, and FMP+BnB~\citep{cai2019practical}.

\begin{itemize}
    \item TOA and Fast-TOA are both tree-oriented approaches and are often used as the baseline methods for evaluating forest point cloud registration methods.
    Fast-TOA is an efficient variant of TOA.
    \RV{
    Using tree positions with DBH values as keypoints, they extract correspondences using the RANSAC strategy coupled with geometric and attribute constraints derived from the keypoints. 
    The correspondences are used to determine a least-squares fit for the rigid transformation.
    }
    \item FMP+BnB is one of the state-of-the-art methods for pairwise registration of general TLS scans~\citep{ge2020object, li2021pairwise}.
    \RV{
    Intrinsic-shape-signatures keypoints~\citep{zhong2009intrinsic} are extracted from the two point clouds, described using fast point feature histograms~\citep{rusu2009fast}, and then matched to generate initial correspondences. 
    Fast match pruning (FMP) is used to prune the correspondence set, followed by fast branch-and-bound (BnB) on the remaining correspondences to determine the optimal registration parameters.
    }
\end{itemize}

Note that TOA and Fast-TOA are not end-to-end solutions: they cannot directly consume raw point clouds but require stem maps (i.e., stem positions and DBHs) as input.
For a fair comparison, we prepared stem maps for every scan as has been done in the Fast-TOA paper~\citep{tremblay2018towards}.
Moreover, TOA was not able to produce any results for scans containing a large number (e.g., greater than 70) of trees. 
Fast-TOA also becomes quite inefficient when handling point clouds consisting of more than 100 trees. For example, it took over an hour to register two scans containing only 117 and 109 trees, respectively.
Hence, we had to reduce the number of trees in every stem map to approximately 50 (similar to the number of trees in the experimental data used in the original paper~\citep{tremblay2018towards}) to be able to conduct the comparison.
To guarantee sufficient overlap between each pair of simplified stem maps, we manually selected these 50 stems to include all corresponding ones.

\subsubsection{Effectiveness}
Following the common practice for plot-scale forest measurement, we used all coarse registration methods to register side scans of each plot to the central ones. 
Since all scan pairs overlap sufficiently, our approach and the three competing methods were able to register them all.
\RV{
All coarse registration results were further refined using the ICP algorithm~\citep{paulj1992method}.
Both the coarse and fine registration errors are reported in Table~\ref{table: coarse results}.
}

Our method outperforms the others in terms of \RV{coarse} registration accuracy on three of the six test plots and achieves the best overall performance with a minimum average pointwise error of 5.9~cm.
\RV{
We can also see that the ICP step significantly improved the coarse registration results, and it eliminated the accuracy gap between coarse registration methods.
This indicates that, when evaluating a coarse registration method, we should prioritize its efficiency and success rate over accuracy.
}

\subsubsection{Efficiency}
\label{run time}
\begin{figure}[t]
    \centering
    \includegraphics[width=0.95\linewidth]{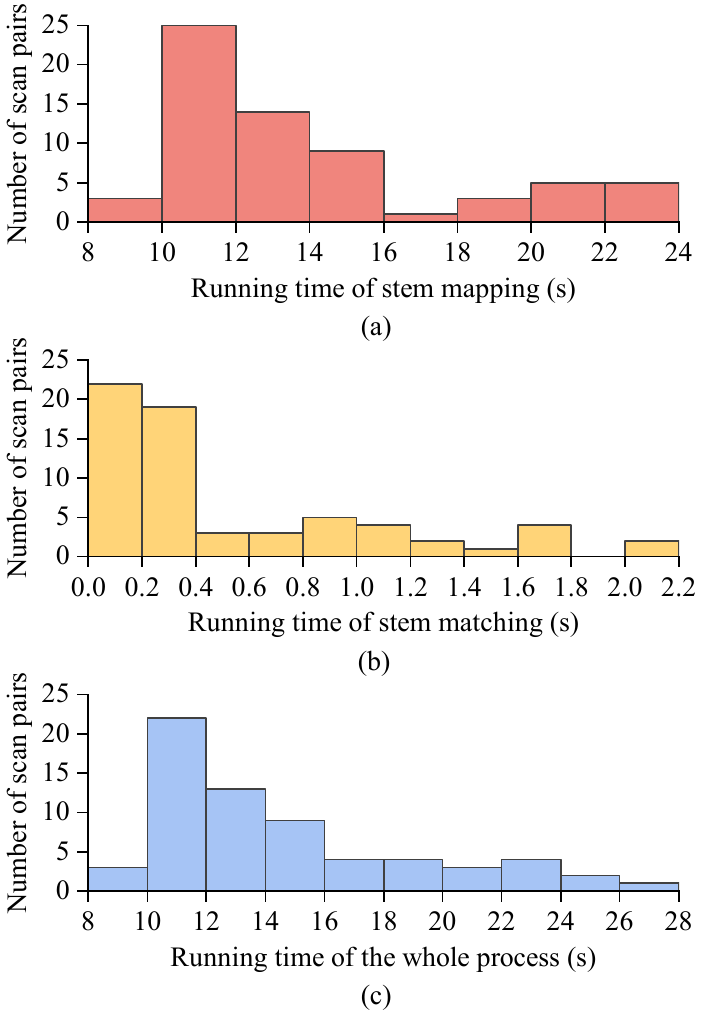}
    \caption{Histograms of running times of the proposed method. 
    (a)~The stem mapping stage. 
    (b)~The stem matching stage. 
    (c)~The whole coarse registration process.}
    \label{fig: efficiency statistics}
\end{figure}

The running times of our method are reported in Fig.~\ref{fig: efficiency statistics}.
The average running times of the stem mapping and stem matching stages are 14.1 seconds and 0.6 seconds, respectively. 
The stem mapping stage takes most of the time, while the stem matching stage is relatively fast, taking less than three seconds even when handling scans containing a large number of trees (more than 200 trees/scan).

\begin{figure}[t]
    \centering
    \includegraphics[width=0.95\linewidth]{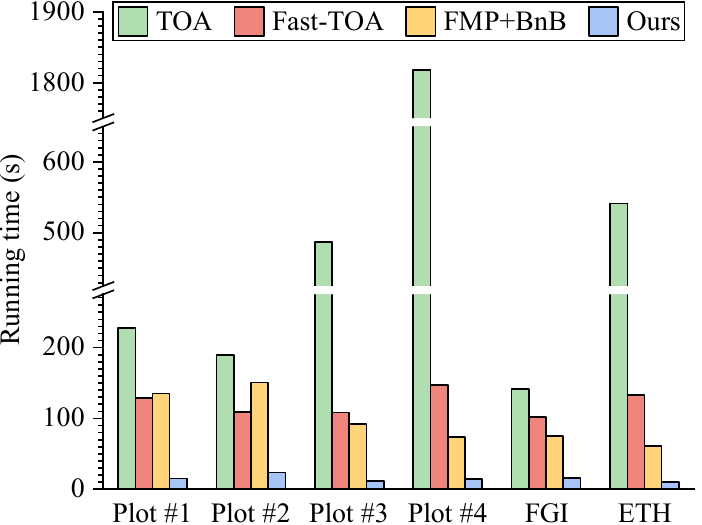}
    \caption{
    \RV{Comparison of running times. 
    Note that the running times of TOA~\citep{kelbe2016marker} and Fast-TOA~\citep{tremblay2018towards} contain only their stem matching part that performs on simplified stem maps ($\sim$50 trees/scan), and the times for the creation of their input (stem maps) were not included. 
    For FMP+BnB~\citep{cai2019practical}, the running time includes both keypoint detection and matching, and similarly for our method, the running time includes both stem mapping and stem matching.}
    }
    \label{figure running time}
\end{figure}

Fig.~\ref{figure running time} reports the running times of our proposed method and the competing methods.
We can see that our approach is much faster than the two stem-based methods, i.e., TOA and Fast-TOA, even though their preparation times of stem maps were not counted and their input was already simplified.
Notably, TOA (relying heavily on DBH values) took unacceptably long times on Plot~\#3, Plot~\#4, and ETH-Trees because trees in these plots have similar attributes.
Our method tackles the inefficiency problem of the stem-based strategy by significantly reducing the complexity of stem matching, which allows it to handle scans containing a large number of trees and leads to consistently high efficiency.
Compared to the local-descriptor-based approach FMP+BnB that intends to accelerate the coarse registration of general TLS scans, the running time of our method is only 1/6 of it on average.

\subsubsection{Robustness}

\begin{figure}[t]
    \centering
    \includegraphics[width=0.95\linewidth]{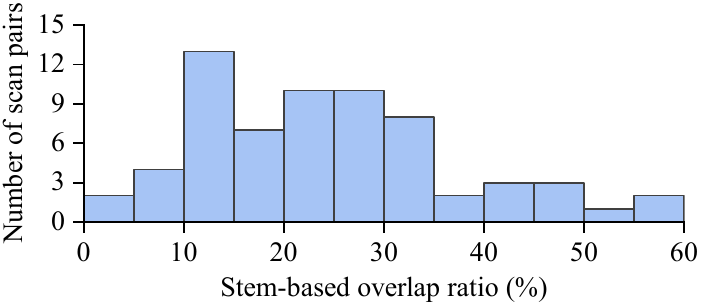}
    \caption{Histogram of stem-based overlap ratios on all scan pairs of the three test datasets.}
    \label{fig: overlap ratio}
\end{figure}

\begin{figure*}[t]
    \centering
    \includegraphics[width=0.95\linewidth]{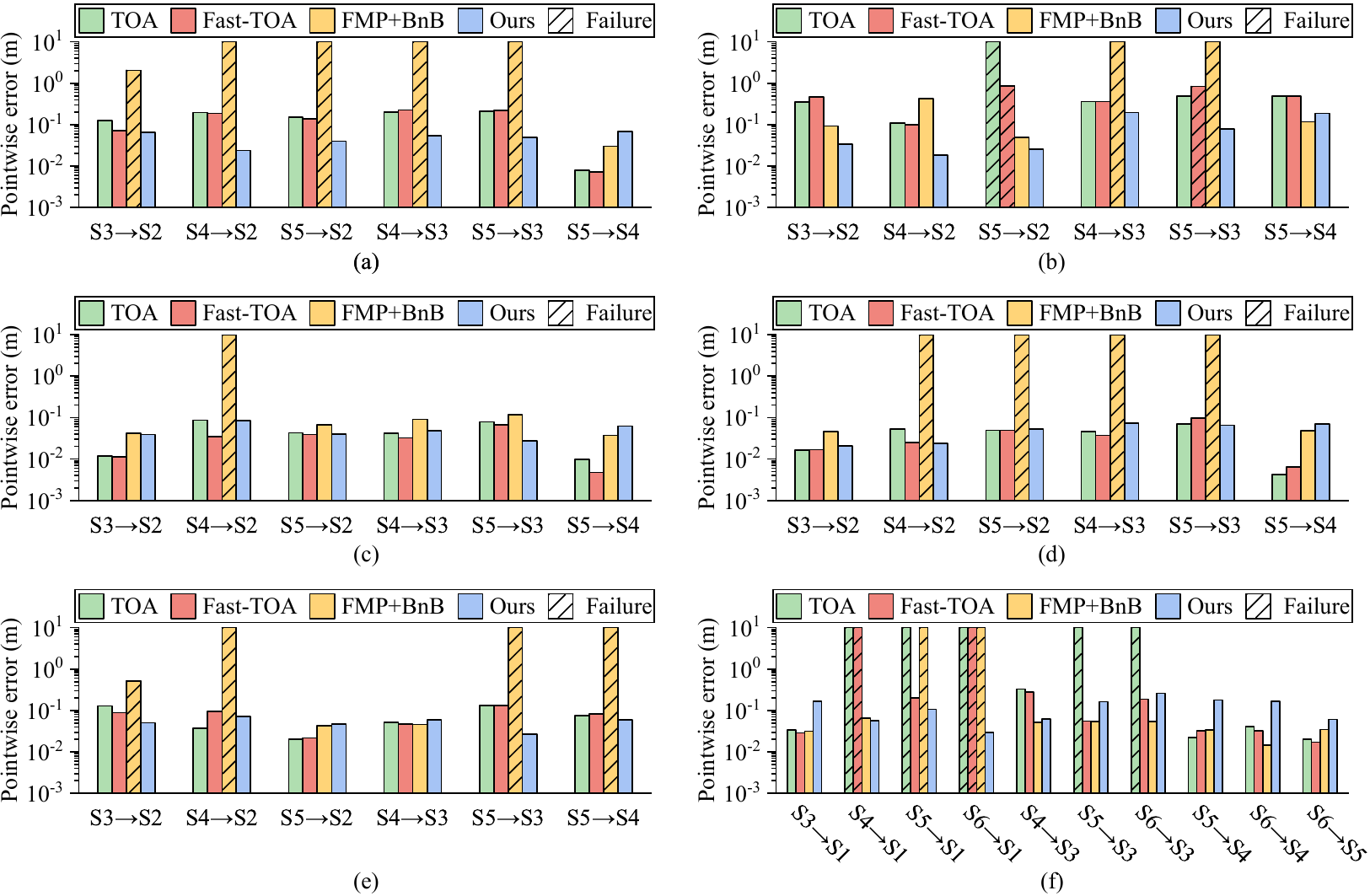}
    \caption{Pointwise errors (in logarithmic scale) for the registration of all other scan pairs in each plot of the three test datasets (in addition to the registration accuracy reported in Table~\ref{table: coarse results}). 
    (a) to (d)~Plots \#1--\#4 of the Tongji-Trees dataset, respectively.
    (e)~WHU-FGI~\citep{dong2020registration}.
    (f)~ETH-Trees~\citep{theiler2015globally}. For the visualization purpose, the error value for any failed registration with a pointwise error exceeding 10~m is truncated to 10~m.}
    \label{fig: robustness comparison}
\end{figure*}

\begin{table*}[t]
\footnotesize
\caption{Success rates of all coarse registration methods}
\label{table: success rate}
\centering
\setlength{\tabcolsep}{10 pt}
\begin{tabular}{c c c c c c c c}
\hline
\multirow{2}{*}{\textbf{Method}} & \multicolumn{7}{c}{\textbf{Success rate of coarse registration (\%)}} \\
                                 & Plot~\#1 & Plot~\#2 & Plot~\#3 & Plot~\#4 & WHU-FGI & ETH-Trees & Average       \\
\hline
TOA                              & 100      & 90       & 100      & 100      & 100 & 67  & 91            \\
Fast-TOA                         & 100      & 80       & 100      & 100      & 100 & 87  & 94            \\
FMP+BnB                          & 50       & 80       & 90       & 60       & 70  & 87  & 74            \\
Ours                             & 100      & 100      & 100      & 100      & 100 & 100 & \textbf{100}  \\
\hline
\end{tabular}
\end{table*}

We have also tried to exhaustively register all scan pairs of each plot to understand the robustness of our method concerning overlap ratio.
The three datasets form a total of 65 scan pairs.
Some of these scan pairs have a small overlap (see Fig. \ref{fig: overlap ratio}), which usually fails marker-free registration methods.
By providing simplified stem maps ($\sim$~50 trees/scan), TOA registered 59 scan pairs while Fast-TOA succeed on 61.
FMP+BnB failed in aligning almost a quarter of scan pairs (with only 48 successful) due to the limited overlap of the input scan pairs.
In contrast, our method managed to register all 65 raw scan pairs.
We have recorded the success rate of all competing methods on the coarse registration of each plot.
The results are reported in Table~\ref{table: success rate}, from which we can see that the forest-oriented registration methods (i.e., TOA, Fast-TOA, and ours) are more robust than the general local descriptor-based methods (i.e., FMP+BnB). 
Among these methods, our method has the highest successful registration rate (i.e., no failure cases).

In addition to the registration errors reported in Table~\ref{table: coarse results}, we also recorded the pointwise errors for the registration of all other scan pairs in each plot of the three test datasets. 
Our method achieves the best performance on 47.5\% (19 out of 40) scan pairs. 
Intuitive visualization of the comparison results is demonstrated in Fig.~\ref{fig: robustness comparison}.

\subsection{Ablation study}

\subsubsection{Stem mapping with/without the verticality constraint}

In the stem mapping stage, we apply a verticality constraint to filter out understory points, which speeds up the subsequent stem modeling process.
We have conducted an ablation study to validate the effects of this constraint.
\RV{
Specifically, we ran our stem mapping algorithm with and without the verticality constraint and evaluated the results based on the ground truth (i.e., manually identified stem positions). 
Table~\ref{Table: ablation verticality} reports the results.
We can see that the verticality constraint accelerated the stem mapping process.
On the other hand, even without the verticality constraint, our method still correctly registered all scan pairs.
The results indicate that our approach is not limited to vertical trees.
}

\begin{table*}[htb]
\footnotesize
\caption{
\RV{Comparison of stem mapping with and without the verticality constraint.
$e_p$ denotes the pointwise error of coarse registration results.}
}
\label{Table: ablation verticality}
\centering
\setlength{\tabcolsep}{10 pt}
\begin{tabular}{c c c c c}
\hline
\multirow{2}{*}{\textbf{Plot ID}} & \multirow{2}{*}{\textbf{Verticality}} & \textbf{Running time} & \textbf{Success rate} & \textbf{$e_p$~(coarse)} \\
                           &        & (s)  & (\%) & (cm) \\
\hline
\multirow{2}{*}{Plot~\#1}  & \cmark & 7.9   & 100  & 6.2  \\
                           & \xmark & 11.8  & 100  & 5.6  \\
\hline
\multirow{2}{*}{Plot~\#2}  & \cmark & 12.2  & 100  & 6.2  \\
                           & \xmark & 94.1  & 100  & 13.4 \\
\hline
\multirow{2}{*}{Plot~\#3}  & \cmark & 5.8   & 100  & 3.6  \\
                           & \xmark & 91.0  & 100  & 4.1 \\
\hline
\multirow{2}{*}{Plot~\#4}  & \cmark & 6.5   & 100  & 3.1  \\
                           & \xmark & 38.4  & 100  & 2.6 \\
\hline
\multirow{2}{*}{WHU-FGI}   & \cmark & 7.7   & 100  & 7.6  \\
                           & \xmark & 102.4 & 100  & 8.1 \\
\hline
\multirow{2}{*}{ETH-Trees} & \cmark & 6.1   & 100  & 8.7  \\
                           & \xmark & 85.6  & 100  & 25.1 \\
\hline
\end{tabular}
\end{table*}

\begin{table}[t]
\footnotesize
\caption{
\RV{
Accuracy comparison of stem matching algorithms implemented in TOA~\citep{kelbe2016marker}, Fast-TOA~\citep{tremblay2018towards}, and ours.
Using simplified stem maps (SSM) as input, all matching methods successfully aligned all scan pairs from the side scans of each plot to the central ones.
$e_p$ denotes the pointwise error, while $e_R$ and $e_t$ represent the two components (i.e., rotation and translation errors) of matrix-based errors, respectively.}
}
\label{table: matching comparison}
\centering
\setlength{\tabcolsep}{3.0 pt}
\begin{tabular}{c c c c c c}
\hline
\multirow{2}{*}{\textbf{Plot ID}} &
  \multirow{2}{*}{\textbf{Method}} &
  \multicolumn{4}{c}{\textbf{Coarse registration errors}} \\
  &                                    & $e_p$ (cm)   & & $e_R$ (mrad)  & $e_t$ (cm)  \\
\hline
\multirow{3}{*}{Plot \#1}  & TOA        & 11.0          & & 10.9      & 8.7     \\
                           & Fast-TOA   & 9.4           & & 9.2       & 6.7     \\
                           & SSM + ours & \textbf{4.4}  & & 0.7       & 4.5     \\
\hline
\multirow{3}{*}{Plot \#2}  & TOA        & 27.8          & & 22.4      & 21.4    \\
                           & Fast-TOA   & 27.3          & & 24.7      & 19.7    \\
                           & SSM + ours & \textbf{20.5} & & 2.0       & 20.3    \\
\hline
\multirow{3}{*}{Plot \#3}  & TOA        & 1.4           & & 1.5       & 1.3     \\
                           & Fast-TOA   & \textbf{1.3}  & & 1.3       & 1.2     \\
                           & SSM + ours & 3.2           & & 0.9       & 3.2     \\
\hline
\multirow{3}{*}{Plot \#4}  & TOA        & 3.9           & & 3.6       & 3.8     \\
                           & Fast-TOA   & 3.7           & & 3.6       & 3.5     \\
                           & SSM + ours & \textbf{1.0}  & & 0.3       & 1.0     \\
\hline
\multirow{3}{*}{WHU-FGI}   & TOA        & 7.6           & & 7.6       & 5.5     \\
                           & Fast-TOA   & 6.4           & & 7.5       & 3.7     \\
                           & SSM + ours & \textbf{5.0}  & & 1.0       & 5.0     \\
\hline
\multirow{3}{*}{ETH-Trees} & TOA        & 20.7          & & 22.8      & 18.2    \\
                           & Fast-TOA   & 22.2          & & 20.0      & 20.0    \\
                           & SSM + ours & \textbf{6.5}  & & 3.5       & 6.0     \\
\hline
\multirow{3}{*}{Average}   & TOA        & 12.1          & & 11.5      & 9.8     \\
                           & Fast-TOA   & 11.7          & & 11.1      & 9.1     \\
                           & SSM + ours & \textbf{6.8}  & & 1.4       & 6.6     \\
\hline
\end{tabular}
\end{table}

\begin{figure}[t]
    \centering 
    \includegraphics[width=0.95\linewidth]{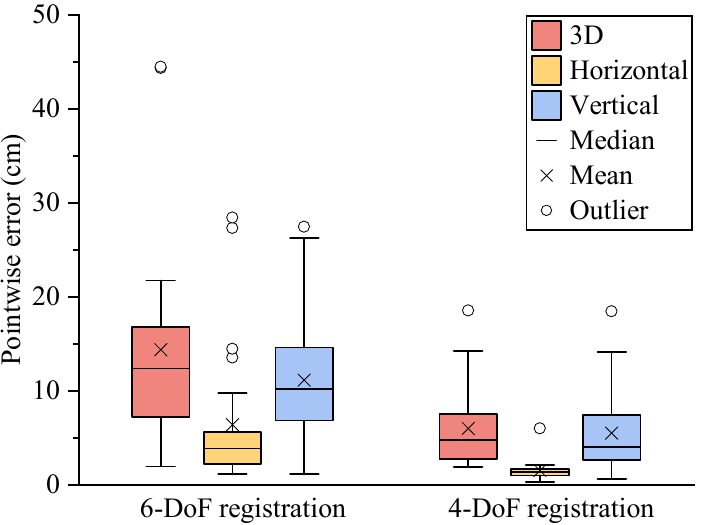}
    \caption{Box plot showing the horizontal and vertical deviations of pointwise registration errors, for both the 6-DoF registration and the 4-DoF registration. 
    ``3D'' denotes the pointwise error, while ``Horizontal'' and ``Vertical'' represent the horizontal and vertical components of the pointwise error, respectively. 
    ``Mean'' and ``Median'' are the respective average and middle values of all pointwise errors. 
    ``Outlier'' is defined as an error value located outside the whiskers of the box plot.}
    \label{fig: boxplot}
\end{figure}

\begin{figure}[t]
    \centering
    \includegraphics[width=0.95\linewidth]{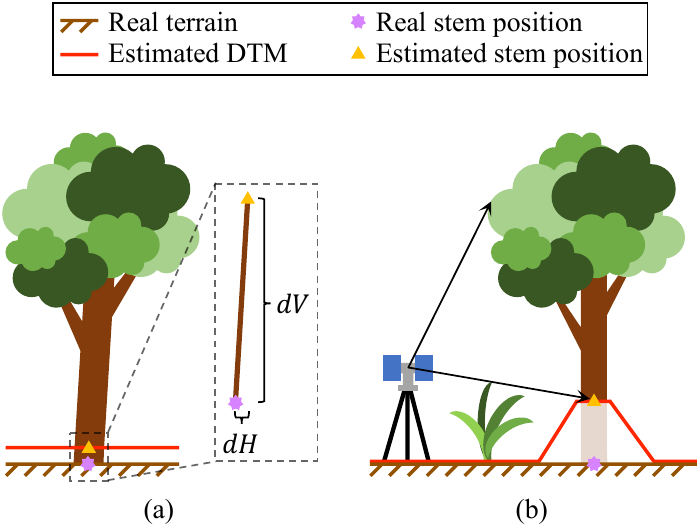}
    \caption{Inaccuracy in the extracted stem positions. 
    (a)~The vertical component of stem positions is sensitive to the change of DTM. 
    (b)~Insufficient ground data (due to occlusions) results in a large error in the $z$ coordinate of the stem position.}
    \label{fig: unbalanced accuracy}
\end{figure}

\subsubsection{\RV{Comparison with stem matching methods}}
\RV{
We performed an ablation study to investigate the efficacy of our matching approach. 
Specifically, we registered side scans of each plot to the central ones by utilizing the tree positions of simplified stem maps (SSM) used by TOA/Fast-TOA algorithms as the input for our matching approach. 
Table~\ref{table: matching comparison} reports the results. 
Our matching approach produced the most accurate results on five out of six test plots, showing the best overall performance with an average pointwise error of 6.8~cm.
}

\subsubsection{6-DoF registration versus 4-DoF registration}
\label{comparison of 6-DoF and 4-DoF}
As has been explained in Section~\ref{method_registration}, our method allows both 6-DoF registration and 4-DoF registration for well-leveled TLS forest scans, and for such data, the 4-DoF solution is preferred.
To understand the effectiveness of both the 6-DoF registration and the 4-DoF registration strategies, we compared them using the three test datasets.
This comparison has revealed that the coarse registration results from both the 6-DoF and the 4-DoF strategies are sufficiently accurate for the subsequent ICP fine registration to converge.
We further computed the horizontal and vertical deviations of the pointwise errors and depicted them in a box plot shown in Fig.~\ref{fig: boxplot}.
From this figure, we can see that the 4-DoF registration results are more accurate than those from the 6-DoF registration, and the 4-DoF registration strategy is remarkably accurate in the horizontal direction (with a mean horizontal pointwise error of only 1.5~cm).
The 4-DoF strategy is more accurate because it takes the advantage of well-leveled scans.

It is also worth noting that the registration errors in the vertical direction are greater than in the horizontal direction. This is a common limitation of the stem-based methods~\citep{liu2017automated, kelbe2016marker, tremblay2018towards, dai2020fast, ge2021global} because stem positions are more accurate in the horizontal direction than those in the vertical direction.
In other words, the $z$ coordinates of the extracted stem positions are more sensitive to the change of DTM, compared with their $x$ and $y$ coordinates, as illustrated in Fig.~\ref{fig: unbalanced accuracy}(a).
The registration results depicted as outliers in Fig.~\ref{fig: boxplot} were obtained for the scans in Plot~\#2 and ETH-Trees where the ground near stems was severely occluded by understory vegetation, leading to large errors in the $z$ coordinates of the stem positions. 
This can be explained from the illustration shown in Fig.~\ref{fig: unbalanced accuracy}(b). 
Such a test indicated that by decomposing the registration problem into two sub-problems, i.e., horizontal alignment and vertical alignment, the 4-DoF strategy can take advantage of the accurate horizontal coordinates of stem positions.

\subsection{Limitation}
\label{limitation}
\begin{figure}[t]
    \centering
    \includegraphics[width=0.95\linewidth]{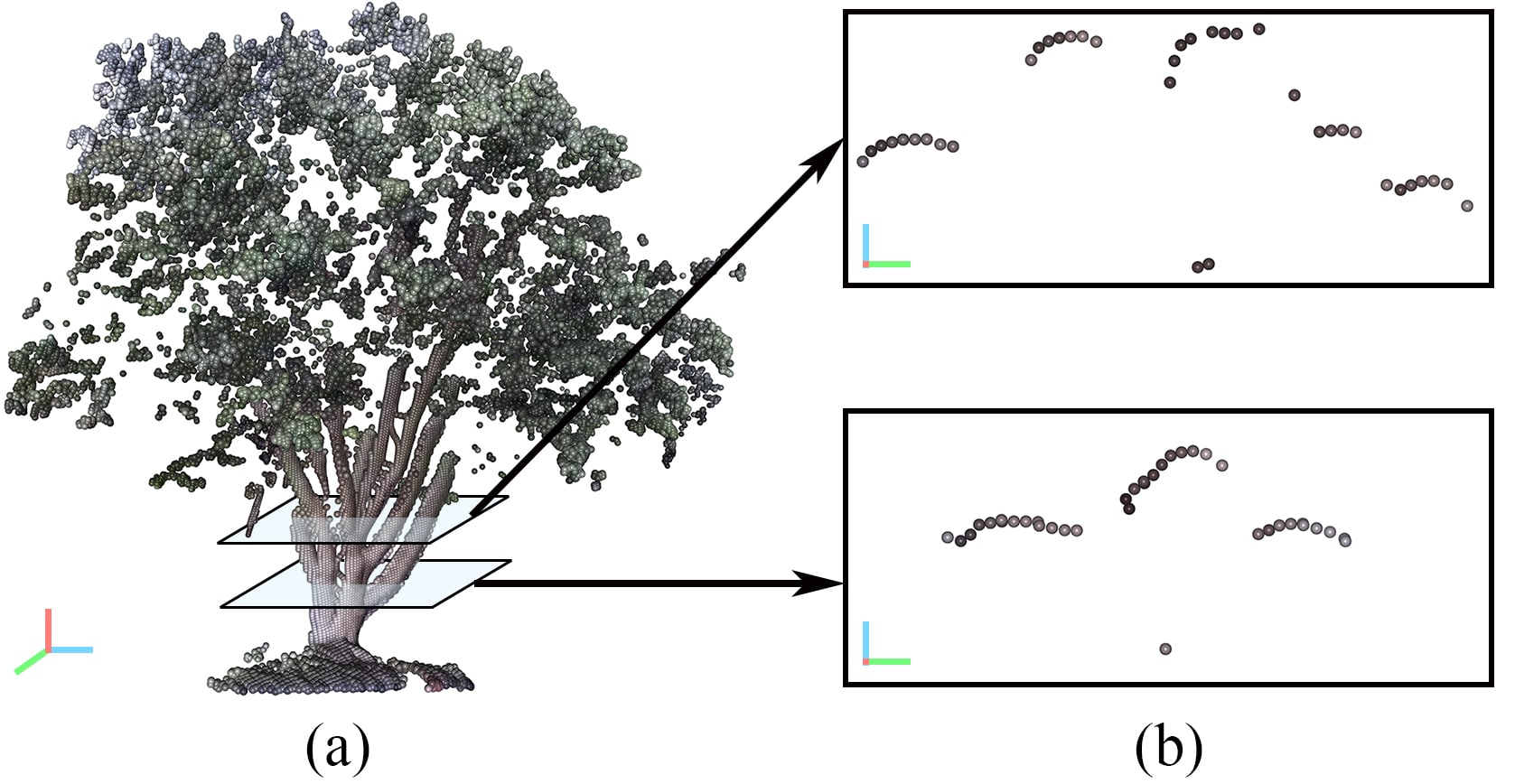}
    \caption{A tree whose stem cannot be modeled using a cylinder. 
    (a)~Point cloud. 
    (b)~Two horizontal cross-sections of the point cloud.}
    \label{figure: limitation}
\end{figure}

Our stem mapping algorithm exploits cylinder primitives to extract tree stems, which may not handle tree stems with complex structures, e.g., tree stems consisting of multiple branches from the bottom (see Fig.~\ref{figure: limitation}). 
Since the large proportion of trees in the tested forest types has a single main stem and the complex stems will be treated as outliers and filtered out in the global matching step, we have not encountered such issues in our experiments.
However, our method will fail if the majority of the trees have such complex structures.

\section{Conclusion}
\label{conclusion}
We have introduced an efficient and robust method to align TLS point clouds captured in forest environments.
Our method accelerates the current stem-based registration strategy in both stages by proposing a fast stem mapping method as well as a robust and efficient stem matching algorithm requiring only the extracted stem positions.
Compared to existing forest-oriented registration techniques, our method performs on par or better regarding effectiveness and robustness while outperforming them significantly in terms of efficiency.
We have also released a new benchmark dataset to the community to enable reliable evaluation and comparison of registration algorithms for forest point clouds.

Our method uses object-level primitives (i.e., tree stems) for registration, which has shown excellent performance. 
In future work, we would like to integrate this approach into frameworks for registration of multiview data, co-registration of multiplatform data, and fusion of spatial-temporal data.

\section*{Acknowledgements}
This work was funded by the National Natural Science Foundation of China (No. 41974213). 
Zexin Yang is supported by the China Scholarship Council.

\bibliographystyle{elsarticle-harv} 
\bibliography{bib/references.bib}
\end{document}